\documentclass[10pt,twocolumn,letterpaper]{article}

\usepackage[pagenumbers]{iccv} % To force page numbers, e.g. for an arXiv version

\usepackage{amsmath, amssymb, algorithm, algpseudocode}
\usepackage{graphicx}
\usepackage{multicol}
\usepackage{multirow}
\usepackage{float}
\usepackage{svg}
\usepackage{mdframed}
\usepackage{amsmath}
\usepackage{algorithm}
\usepackage{algpseudocode}
\usepackage{tabularx}   % nice tables
\usepackage{multirow}
\usepackage{pifont}        % For check and cross marks
\usepackage{colortbl}      % Additional table color functionality 
\usepackage{tikz}
\usetikzlibrary{shapes.geometric, calc}
\usepackage{wrapfig}
\usepackage{siunitx}
\usepackage{comment}
\usepackage{array}
\newcolumntype{Y}{>{\raggedright\arraybackslash}X}

\DeclareRobustCommand{\circled}[2]{%
  \tikz[baseline=(char.base)]{
    \node[shape=circle, fill=#2, draw=black, inner sep=0.5pt] (char) {#1};
  }%
}

\newcommand{\inparagraph}[1]{\smallskip\noindent\textbf{#1}\hspace{0.5em}}
\newcommand{\inparagraphnospace}[1]{\smallskip\noindent\textbf{#1}}

\definecolor{orange}{RGB}{255, 128, 41}
\definecolor{pink}{RGB}{229, 158, 221}
\definecolor{yellow}{RGB}{249, 249, 165}
\definecolor{checkyes}{rgb}{1.0, 1.0, 1.0}
\definecolor{crossno}{rgb}{1.0, 1.0, 1.0}
\newcommand{\cmark}{\text{\ding{51}}}%
\newcommand{\xmark}{\text{\ding{55}}}%
\def \y {$\cmark$\cellcolor{checkyes}}
\def \n {$\xmark$\cellcolor{crossno}}

\usepackage{tabularx}
\newcolumntype{C}{>{\centering\arraybackslash}X}

\usepackage{siunitx} % table alignment at decimal point
\newrobustcmd\B{\DeclareFontSeriesDefault[rm]{bf}{b}\bfseries}

\usepackage{makecell}

\usepackage{xr-hyper}

\makeatletter
\newcommand*{\addFileDependency}[1]{
  \typeout{(#1)}
  \@addtofilelist{#1}
  \IfFileExists{#1}{}{\typeout{No file #1.}}
}
\newcommand*{\newbibstartnumber}[1]{%
  \apptocmd{\thebibliography}{%
    \global\c@NAT@ctr #1\relax
    \addtocounter{NAT@ctr}{-1}%
  }{}{}%
}
\makeatother

\definecolor{iccvblue}{rgb}{0.21,0.49,0.74}
\usepackage[pagebackref,breaklinks,colorlinks,allcolors=iccvblue]{hyperref}

\def\paperID{12811} % *** Enter the Paper ID here
\def\confName{ICCV}
\def\confYear{2025}

\title{ART: Adaptive Relation Tuning for Generalized Relation Prediction}

\author{%
  \begin{minipage}[t]{\textwidth}
    \centering
    {\large
      Gopika Sudhakaran\textsuperscript{1,2} \quad
      Hikaru Shindo\textsuperscript{1} \quad
      Patrick Schramowski\textsuperscript{1,3} \quad
      Simone Schaub-Meyer\textsuperscript{1,2} \quad
      Kristian Kersting\textsuperscript{1,2,3} \quad
      Stefan Roth\textsuperscript{1,2} \\[0.5em]
    }
    {\normalsize
    \textsuperscript{1}Department of Computer Science, TU Darmstadt, Germany\\
    \textsuperscript{2}Hessian Center for AI (hessian.AI) \quad
    \textsuperscript{3}German Research Center for AI (DFKI)\\
    {\color{RoyalBlue}\url{https://github.com/visinf/ART}}}\\[1em]
  \end{minipage}
}

\usepackage{fancyhdr}
\usepackage{setspace}

\fancypagestyle{firststyle}{

  \fancyhf{}
  \lfoot{
    {\footnotesize
    \begin{spacing}{0.5}
    \parbox{\linewidth}{
      \vspace{-0.85em}
      To appear in \emph{Proceedings of the IEEE/CVF International Conference on Computer Vision (ICCV)}, Honolulu, Hawai'i, USA, 2025.\\\hrule\vspace{\baselineskip}
      \copyright~2025 IEEE. Personal use of this material is permitted. Permission from IEEE must be obtained for all other uses, in any current or future media, including reprinting/republishing this material for advertising or promotional purposes, creating new collective works, for resale or redistribution to servers or lists, or reuse of any copyrighted component of this work in other works.
    }
    \end{spacing}
    }
  }
}

\begin{document}
\maketitle
\begin{abstract}
Visual relation detection (VRD) is the task of identifying the relationships between objects in a scene. VRD models trained solely on relation detection data struggle to generalize beyond the relations on which they are trained. While prompt tuning has been used to adapt vision-language models (VLMs) for VRD, it uses handcrafted prompts and struggles with novel or complex relations. We argue that instruction tuning offers a more effective solution by fine-tuning VLMs on diverse instructional data. 
We thus introduce ART, an \textbf{A}daptive \textbf{R}elation \textbf{T}uning framework that adapts VLMs for  VRD through instruction tuning and strategic instance selection. By converting VRD datasets into an instruction-tuning format and employing an adaptive sampling algorithm, ART directs the VLM to focus on informative relations while maintaining generalizability. Specifically, we focus on the relation classification, where subject-object boxes are given and the model predicts the predicate between them. We tune on a held-in set and evaluate across multiple held-out datasets of varying complexity. Our approach strongly improves over its baselines and can infer unseen relation concepts, a capability absent in mainstream VRD methods. We demonstrate ART's practical value by using the predicted relations for segmenting complex scenes.
\end{abstract}
\thispagestyle{firststyle}
\section{Introduction}
\label{sec:intro}

\begin{figure}[ht]
    \centering
    \includegraphics[width=0.99\linewidth]{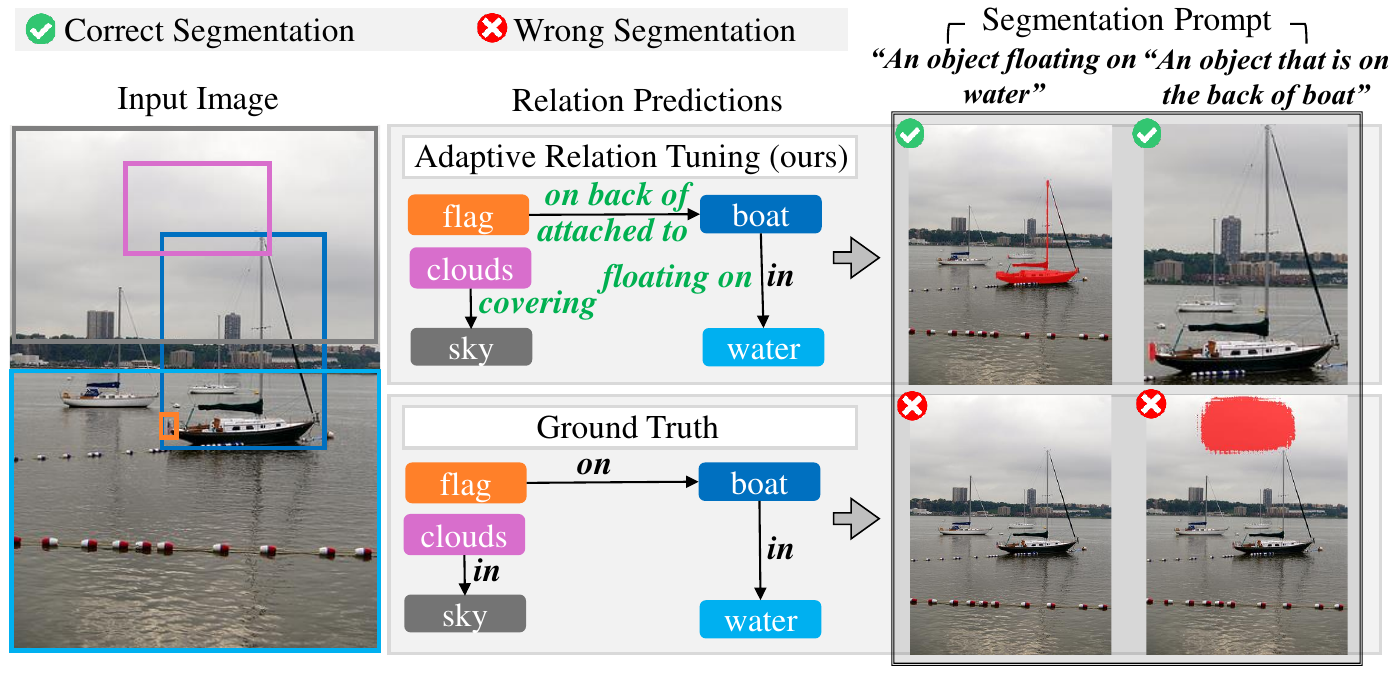}
    \vspace{-0.75em}
    \caption{ 
     \textbf{ART predicts detailed, context-rich relationships, enhancing downstream reasoning, \eg, for segmentation.}  From the input image, ART (trained on VG) predicts unseen and informative relations like \emph{floating on} and \emph{on back of} (highlighted in green). This richer relational context, facilitated by careful tuning of a VLM, \eg, allows the DeiSAM~\cite{shindo2024deisam} segmentation model, which relies on scene graphs for spatial reasoning, to accurately identify even small objects like the flag on the boat's back.  When prompted to find an object floating on water, the ART scene graph also enables correct segmentation, while the less specific ground-truth scene graphs~\cite{hudson2019gqa} may miss such segmentation.}
    \label{fig:floating_boat}
    \vspace{-0.75em}
\end{figure}

Visual relation detection (VRD) plays a key role in visual scene understanding by enabling the recognition of relationships between entities as triplets of $\langle$subject, predicate, object$\rangle$ \cite{zellers2018neural, tang2019learning, he2022towards, li2024zero}.
This structured understanding of visual content is essential for downstream tasks like visual question answering (VQA)~\cite{antol2015vqa}, image captioning~\cite{hossain2019comprehensive}, and referring object detection and segmentation~\cite{shindo2024deisam}. While mainstream VRD models have advanced through relation-centric training, they predominantly rely only on fixed datasets such as Visual Genome~\cite{krishna2017visual} and GQA~\cite{hudson2019gqa}. This dependence introduces three key limitations: \emph{(i)} Overfitting to frequent in-distribution relations, weakening performance on rare relations~\cite{chang2021comprehensive}; \emph{(ii)} Inability to infer novel, unseen relations; \emph{(iii)} Coarse annotations that fail to capture fine-grained relational semantics. For example, given the scene in \cref{fig:floating_boat}, GQA annotations describe generic relations like \emph{``boat in water"} or \emph{``flag on boat"}. A model trained solely on these overlooks finer distinctions, such as whether the boat is \emph{floating on water} or the flag is positioned \emph{on the back of the boat}. Humans naturally infer such nuances using broad context, yet, manually extending VRD datasets with fine-grained annotations is costly and impractical, making it crucial to develop models that generalize beyond predefined relations without exhaustive retraining.

A promising alternative is to leverage vision-language models (VLMs), which excel in generalization across multimodal tasks by learning from large-scale image-text corpora~\citep{radford2021CLIP,flamingo_neurips,liu2023llavainteractive,dai2024instructblip}. Thus, VLMs could help overcome VRD generalization challenges. Although recent studies have attempted to extract knowledge from VLMs to enhance VRD~\cite{he2022towards, li2024zero}, they rely on off-the-shelf VLMs through prompt engineering, crafting prompts tailored to the cues of an in-distribution or otherwise limited set of relations.

To address these challenges, we reframe the VRD task as an instruction tuning (IT) problem, leveraging the proven ability to enhance VLMs~\cite{wei2021finetuned, dai2024instructblip}. We first generate instruction-tuning instances from established VRD benchmarks, followed by fine-tuning with these tailored instances. We propose a novel framework for adaptively selecting the most informative instances for relation tuning. Our approach, named Adaptive Relation Tuning (ART), enhances the relation classification capabilities of VLMs by focusing on the most informative aspects of the data. This is crucial for safety-critical scenarios like autonomous driving, where distinguishing subtle relational differences, \eg, ``pedestrian waiting at crosswalk” versus ``pedestrian stepping onto crosswalk,” can be life-saving. By enabling the detection of fine-grained relations while preserving generalization, ART moves beyond simplistic relational inference, enabling more robust and adaptable visual reasoning.

Our key contributions are: \emph{(i)} 
We convert benchmark relation detection datasets into an effective format for instruction tuning.
\emph{(ii)} We present ART, an innovative framework for relation tuning of VLMs, which preserves the model's generalization by adaptively selecting informative samples. 
\emph{(iii)} We train ART on a held-in dataset and evaluate it on held-out datasets of varying complexity. Our quantitative analysis highlights ART's strong generalization capabilities.
\emph{(iv)} We deploy ART on a downstream segmentation task, where both quantitative and qualitative results highlight its real-world effectiveness.
\section{Related work}
\textbf{Visual relation detection (VRD).} The prediction  of relationships between subject-object pairs has been widely studied in the domain of scene graph generation (SGG)~\cite{lu2016visual, liao2019natural, zellers2018neural, tang2020unbiased} and human object interaction (HOI)~\cite{gao2018ican, ulutan2020vsgnet, iftekhar2023gtnet, xu2019learning}. Previous work largely focused on learning the object and predicate categories in the training data distribution and testing on the same distribution. However, such models suffer from noisy annotations~\cite{li2022devil} and biased predictions resulting from the long-tailed predicate distribution~\cite{tang2020unbiased, zheng2023prototype, sudhakaran2023vision}. Thus, current mainstream models fall short in their ability to generalize beyond the specific relations they were trained on, highlighting a critical gap in the field. This underscores the need for innovative approaches that can enhance model robustness and generalization capabilities, enabling effective inference over completely unseen object classes and relationships. Recently, some efforts were made in this direction  by adopting prompt tuning~\cite{liu2023pre} of VLMs~\cite{he2022towards, li2024zero}. However, these approaches rely on off-the-shelf VLMs with prompts tailored to limited relational cues. Conversely, instruction tuning~\cite{dai2024instructblip} offers a more comprehensive solution by fine-tuning models on diverse instructional data, significantly enhancing their ability to generalize and infer unseen relationships. Hence, we here adapt VLMs for VRD with adaptive instruction tuning.
%enhance the relation prediction capability of VLMs with adaptive instruction tuning.

\begin{comment}
\begin{figure}[t]
    \centering
    % \includegraphics[width=.95\linewidth]{figures/deisam_result_new.pdf}
    \includegraphics[width=.95\linewidth]{figures/segmentation_new.pdf}
    \vspace{-0.5em}
        \caption{\textbf{ART can be used to label missing annotations and predict new unseen predicates.} Segmentation results with textual prompts \emph{(top)} using DeiSAM~\cite{shindo2024deisam}, which segments objects via reasoning on scene graphs. ART successfully detects new relations and improves the segmentation quality, while ground-truth scene graphs fail to capture relations in the prompt.}
    \label{fig:deisam_result}
    \vspace{-0.5em}
\end{figure}
\end{comment}

\inparagraphnospace{Uncertainty estimation} is essential in setups that require informative sample selection. At a high level, uncertainty can be classified into aleatoric (data) and epistemic (model) uncertainty.  Aleatoric uncertainty reflects intrinsic variability in the data, while epistemic uncertainty arises from the model's limited understanding of the underlying task. Some of the commonly used uncertainty estimators are Monte-Carlo dropout~\cite{gal2016dropout}, deep ensembles~\cite{fort2019deep}, and the entropy of the softmax predictions~\cite{nguyen2022measure}. These uncertainty measures are used predominantly to compute model uncertainty. In our work, we use entropy as an uncertainty measure due to its cost-effective use in a VLM setup compared to dropout and ensembles. We also use a cosine similarity-based approach~\cite{li2013distance} as an additional uncertainty measure. 

\inparagraphnospace{Active learning (AL)} iteratively selects and annotates the most informative instances from an unlabeled pool to improve training~\cite{olsson2009literature, siddhant2018deep, zhang2022survey}. Kung \etal~\cite{kung2023active} proposed selecting informative instructions using an active learning-inspired setup. Our work is related to these techniques but differs in a key aspect: we assume the pool data is already labeled. Our goal is to identify the most informative samples within the provided distribution to achieve optimal performance in out-of-distribution scenarios, while AL aims for optimal in-distribution performance with limited data.
\begin{figure*}[t]
\begin{mdframed}[backgroundcolor=gray!20]
\small

\textbf{Question:} Is there a prominent $\langle r \rangle$ relation between $\langle s \rangle$ (subject) and $\langle o \rangle$ (object) in the image?

\textbf{Positive response:} Yes, $\langle s, r, o \rangle$.

\textbf{Negative response:} No, there is no prominent $\langle nr \rangle$ relation between $\langle s \rangle$ and $\langle o \rangle$. 

\end{mdframed}
\vspace{-1.5em}
\caption{\textbf{Relation tuning format template.}}
\label{fig:prompt_format}
\vspace{-0.5em}
\end{figure*}

\section{Adaptive relation tuning framework}
\label{instruct_art}

VRD models must generalize well to be effective in real-world scenarios. Yet, models trained solely on VRD datasets either overfit the underlying distribution or learn spurious correlations.
 Though VLMs excel in generalization due to their large training corpora, they often struggle with fine-grained relations, favoring frequent patterns. To address this, ART follows a two-step process: \emph{(1)} Relation-tuning data creation, where the training data is structured with high-level relation categories (spatial, semantic, possessive) and a counter-negative mining strategy to refine negative samples; and \emph{(2)} Adaptive relation tuning of VLMs, which optimizes the model to focus on rare, diverse relations, improving generalization to unseen scenarios. 

Typically, VRD involves object detection followed by relation classification.  While generalization in object detection—such as zero-shot or cross-domain detection—has been widely studied~\cite{yan2022semantics, rahman2018zero, yan2020semantics}, we focus on relation classification.  Given object categories and bounding boxes, we predict the relation between subject-object pairs, denoted as $s$ (subject), $o$ (object), and $p$ (predicate/relation). 

\subsection{Relation-tuning data creation}
\label{sec:art:data}

A key aspect of our approach is relation-tuning data construction, which involves carefully crafting questions to capture fine-grained object relations. A simple transformation of the ground-truth relation triplets $\langle s, p, o \rangle$ into a question format such as ``Is there a relation between $\langle s \rangle$ and $\langle o \rangle$ in the image?"—with the answers ``Yes, $\langle s, p, o \rangle$" or ``No" for valid or invalid relations, respectively, might seem sufficient. However, this naive approach overlooks important nuances in relation classification, as shown below (\cref{tab:instruction_set_ablation} in \cref{subsec:exp_results}). To address this, we introduce a more structured approach to relation-tuning data creation, where we design a question format that incorporates two key components: \emph{(1)} high-level relation categories and \emph{(2)} counter-negative mining. The final relation tuning format template, shown in \cref{fig:prompt_format}, is derived from this structured formulation, which is empirically supported by findings in \cref{tab:instruction_set_ablation}.

Prevalent VRD relations can be classified into high-level relation types~\cite{zellers2018neural} such as spatial (\emph{above, behind, under, \ldots}, denoting spatial arrangements of objects), semantic (\emph{carrying, eating, using, \ldots}, corresponding to activities), and possessive (\emph{wearing, part of, has, \ldots}, depicting a sense of ownership). These high-level relation types can help the model learn a richer set of relations. Furthermore, because multiple relations can co-exist between subject-object pairs, we carefully construct questions to capture these complexities. \Eg, the phrase ``person walking on street" has both semantic (action) and spatial (location) relations, and a broad question format might fail to capture possible predictions.

Another challenge arises from the incompleteness of VRD benchmark annotations, where many potential object relations remain unlabelled. Treating all unlabelled pairs as negatives can mislead the model and degrade performance. To address this, we introduce a counter-negative mining approach that selectively assigns negatives based on the relation type. We assume that \emph{semantic \vs possessive} and \emph{spatial \vs possessive} relations are mutually exclusive—semantic and spatial relations serve as negatives for possessive ones, and vice versa.  \Eg, in \emph{person wearing hat} (possessive), spatial orientation is implicit, while action verbs are less relevant, as possession describes a static state rather than an active interaction. 
In contrast, semantic relations often depend on spatial positioning, which can vary and impact their meaning. \Eg, in \emph{person watching TV}, the spatial position of the subject relative to the object is crucial—whether they are near, far, or beside the TV changes the nature of the interaction. Thus, \emph{semantic \vs spatial} is not treated as mutually exclusive.
Specifically, we define the following exclusivity for high-level relation categories, grounded in the above assumptions and empirically validated through our instruction set analysis in \cref{tab:instruction_set_ablation}:
\begin{align}
\text{positive: } r &\in \{\text{spatial}, \text{possessive}, \text{semantic}\}, \\
\text{negative: } nr &= 
\begin{cases} 
\{\text{possessive}\}, & \text{if } r = \text{spatial} ,\\ 
\{\text{spatial}, \text{semantic}\}, & \text{if } r = \text{possessive} ,\\\notag 
\{\text{possessive}\}, & \text{if } r = \text{semantic},
\end{cases}
\end{align}
where $r$ and $nr$ denote positive resp.\ negative categories.

\begin{figure*}[t]%
\centering
\includegraphics[width=0.99\linewidth]{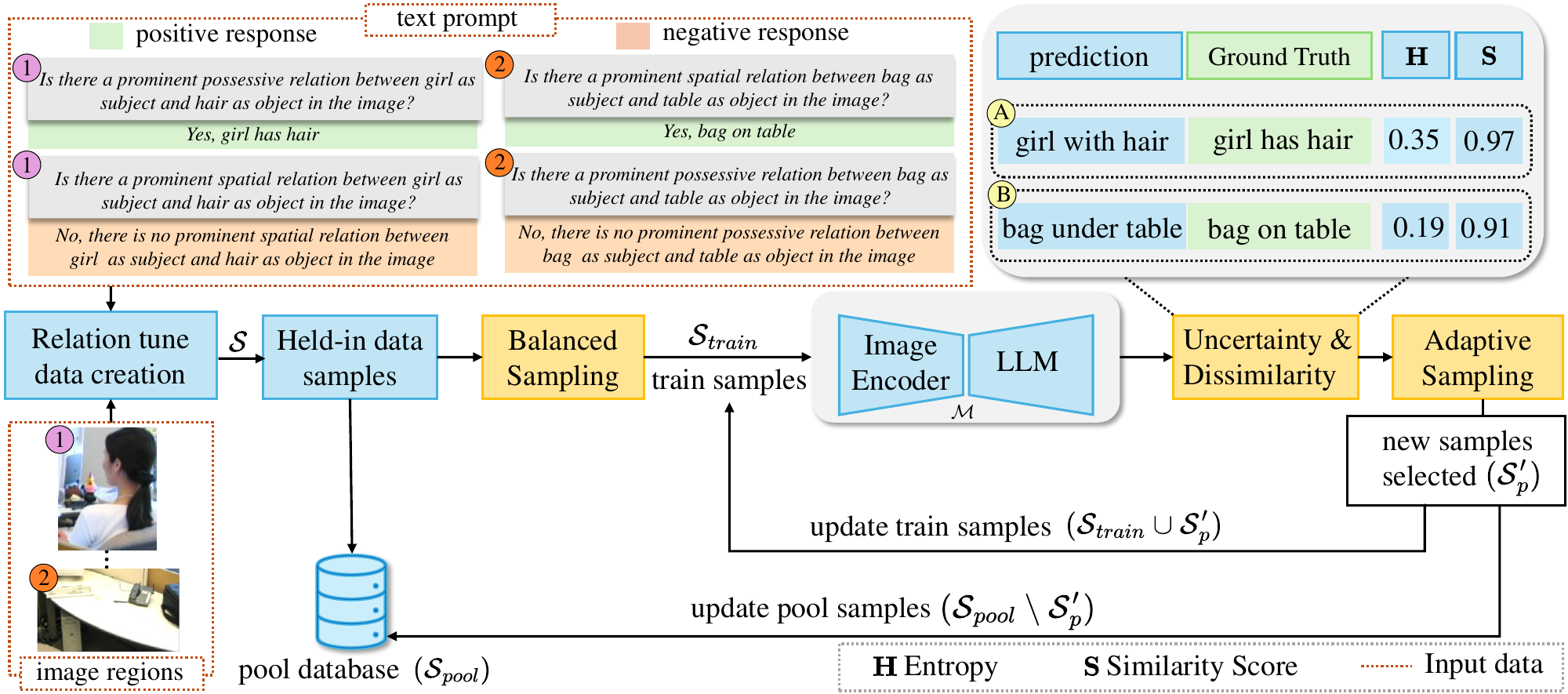}
%\decoRule
\vspace{-0.75em}
\caption{\textbf{ART Framework.} To construct the relation-tuning dataset $\mathcal{S}$, we create question-response prompts for each image region (\eg, \circled{1}{pink} and \circled{2}{orange}), including both positive and negative examples, which we call instruction sets. Balanced sampling ensures that instruction sets span all predicate categories in $\mathcal{S}$, forming the training data $\mathcal{S}_{\text{train}}$. This training data is used to fine-tune the vision-language model (VLM), $\mathcal{M}$, while unused data is stored in the pool $\mathcal{S}_{\text{pool}}$. After training, $\mathcal{M}$ is evaluated on $\mathcal{S}_{\text{pool}}$, where uncertainty and dissimilarity are estimated using entropy $\mathbf{H}$ and similarity scores $\mathbf{S}$. In example \circled{A}{yellow}, the model assigns higher entropy to a similar but uncertain prediction compared to an incorrect prediction \circled{B}{yellow}, but similarity scores help prioritize the incorrect example for further training. These estimates guide adaptive sampling, creating a refined set of samples $\mathcal{S}_{p}'$ for each predicate $p$ to expand $\mathcal{S}_{\text{train}}$, enabling further rounds of training for $\mathcal{M}$.}
\label{fig:art}
\vspace{-0.5em}
\end{figure*}

\subsection{Adaptive relation tuning (ART)}
\label{sec:art}

Given that we have converted a benchmark VRD dataset into instruction-tuning format (\cf \cref{sec:art:data}), a straightforward approach would be to tune the VLM on the full instruction set. 
However, this leads to overfitting and loss of generalization (\cf \cref{tab:ablation_art} below). To prevent this, we propose adaptive relation tuning (ART), which selectively refines structurally relevant relational patterns while maintaining the VLM’s broad generalization for VRD.

Inspired by active learning, ART aims to select the most informative samples through multiple learning iterations, as illustrated in \cref{fig:art}. While both prioritize uncertain instances, ART differs by operating on a fully labeled dataset rather than an unlabeled pool. We aim to achieve optimal out-of-distribution performance without overfitting to in-distribution data. As depicted in \cref{fig:art}, ART comprises three key components (highlighted in orange): a balanced sampling module,  an adaptive sampling module, and an uncertainty and dissimilarity computation module. %Let us explain each.

\inparagraph{Balanced sampling module.}  ART operates through multiple learning iterations, selecting a fixed budget $B$ of training samples per loop to refine the model. However, VRD datasets, such as Visual Genome (VG) \cite{krishna2017visual}, are highly imbalanced, dominated by a few frequent classes. If the initial training samples were randomly selected, the model would be biased toward these head classes, limiting its ability to learn rare relations. To counteract this, our balanced sampling module (see Algorithm \ref{alg:balanced_sampling} in Appendix \ref*{sec:balanced sampling}) serves as the initialization step, ensuring an even distribution of relations at the start of training.  Let $\mathcal{P}=\left\{p_i\right\}_{i=1}^{N}$ represent the set of predicates, where $N$ is the number of predicate categories, and let $B'_p$ and $N_p$ represent the sampling budget and available data samples per predicate.
$B'_p$ is evenly assigned across predicates, ensuring all are included. If $N_p$ is exhausted for a predicate $p$, sampling continues from the remaining predicates until $B$ is met. This prevents early overfitting to frequent relations, allowing adaptive sampling to refine predicate selections in later iterations.

\inparagraph{Adaptive sampling module.} 
Only using balanced sampling can still miss out on informative samples, as the model quickly learns some relations while struggling with others.  To address this, we propose \emph{adaptive sampling}, which prioritizes samples where the VRD model lacks confidence. 
This approach selects the necessary samples from each category based on how well it performs and how uncertain the model is about the samples in each category.

\inparagraph{Uncertainty and dissimilarity computation module.}
Inspired by uncertainty-based active learning~\cite{nguyen2022measure}, we leverage entropy to measure the uncertainty of the VRD model's output.
To achieve this, the output from the LLM component of the VLM is further processed. Since our VLMs employ beam search, we assess uncertainty across multiple possible sequences.  Let $M$ be the beam width, a hyperparameter controlling the number of candidate sequences.  To comprehensively understand the model's uncertainty, we compute the entropy over all possible sequences. Let \(\mathbf{E} \in \mathbb{R}^{M \times L \times V}\) represent the output logits from the LLM, where \(L\) is the sequence length and \(V\) is the vocabulary size. We calculate the overall entropy \(\mathbf{H}\) of the predicted sentence as \(\mathbf{H} = -\frac{1}{M\cdot L} \sum_{m,l,v} \mathbf{P}_{mlv} \log(\mathbf{P}_{mlv})\), where \(\mathbf{P}_{mlv} = \frac{\exp(\mathbf{E}_{mlv})}{\sum_{v'} \exp(\mathbf{E}_{mlv'})}\) is the softmax probability for each vocabulary element \(v\).
 This formula averages the entropy across all beams and sequence positions, providing a single entropy value that summarizes the uncertainty regarding the words in the predicted sentence.

As illustrated in \cref{fig:art} (\cf \circled{A}{yellow} and \circled{B}{yellow}), while entropy can indicate model-level uncertainty, it may also produce low entropy values for overly confident predictions. Relying solely on entropy could result in neglecting informative samples. Since our objective is to develop a generalizable VRD model capable of handling out-of-distribution data, retaining predictions that semantically align with the ground truth is preferable.
For instance, the phrases \emph{girl with hair} and \emph{girl has hair} are semantically similar; hence, such False Positives can be considered valid and sampled less often. Conversely, the prediction \emph{bag under table} for the ground truth \emph{bag on table} is evidently invalid, but the model’s overconfidence may assign it low entropy, overlooking valuable samples. To overcome this issue, we  use  a similarity metric $\mathbf{S}$, measuring the cosine similarity between predicted and ground-truth embeddings, ensuring better selection of informative samples.

\inparagraph{Adaptive sampling algorithm.} We summarize the adaptive sampling algorithm in \cref{alg:adaptive_sampling}. Given a fully labeled pool $\mathcal{S}_{\text{pool}}$, we begin by categorizing the samples into True Positives (TP), False Positives (FP), and False Negatives (FN) by inference on $\mathcal{S}_{\text{pool}}$.
A TP occurs when the predicted positive response matches the ground truth. An FP arises when the predicted positive response does not align with the ground truth. An FN happens when the model predicts a negative response for a relation that actually exists in the ground truth.   This categorization is essential for identifying areas where the model can be improved. Subsequently, our approach adaptively adjusts sampling thresholds based on entropy and similarity scores of $\mathcal{S}_{\text{pool}}$, ensuring appropriate sample selection at each iteration. The following steps outline the sampling algorithm in detail.

\textbf{Step 1: Sampling budget allocation.} To prioritize predicates requiring greater supervision, we compute a per-predicate sampling probability as $P_p = \frac{1 - R_p}{\sum_{j} (1 - R_j)}$, where $R_p$ is the recall of predicate $p$ on the validation set. Each predicate is allocated a budget $B'_p = \min(B \cdot P_p, N_p)$, ensuring that predicates with lower recall receive a higher sampling budget while preventing over-sampling beyond available instances (Lines 7--10). 

\textbf{Step 2: Sampling process.} We iteratively select informative samples by leveraging predicate-specific uncertainty and dissimilarity. This process consists of three key components: \emph{(i)} \emph{Adaptive thresholding}, where thresholds are adapted based on entropy $\mathbf{H}$ and similarity $\mathbf{S}$ distributions; \emph{(ii)} \emph{Sampling criteria}, which define selection rules for TPs, FNs, and FPs based on their statistics; and \emph{(iii)} \emph{Threshold refinement}, ensuring sufficient sample selection when the allocated budget is not met. 

\begin{algorithm} [t]
\caption{Adaptive Sampling Algorithm}
\label{alg:adaptive_sampling}
\begin{algorithmic}[1]

\State \textbf{Input:}  
\State \hspace{1em} $\mathcal{S}_{\text{pool}} \gets \mathcal{S} \setminus \mathcal{S}_{\text{train}}$ 
\State \hspace{1em} $B \gets$ Total sampling budget per iteration
\State \hspace{1em} $\mathcal{R} \gets$ Recall per predicate on validation set $\mathcal{S}_{\text{val}}$
\State \hspace{1em} $\text{TP}_{\text{pool}}, \text{FN}_{\text{pool}}, \text{FP}_{\text{pool}}$ \Comment{True Positive, False Negative, and False Positive samples in $\mathcal{S}_{\text{pool}}$}

\State \textbf{Output:} Updated training set $\mathcal{S}_{\text{train}}$ and pool set $\mathcal{S}_{\text{pool}}$

\State \textbf{Step 1: Sampling Budget Allocation}
\For{each predicate $p$}
    \State $P_p \gets \frac{1 - R_p}{\sum_{j} (1 - R_j)}$ \Comment{Higher weight for lower recall}
    \State $B'_p \gets \min(B \cdot P_p, N_p)$ \Comment{$N_p$: available data samples for $p$}
\EndFor

\State \textbf{Step 2: Sampling Process}
\For{each predicate $p$ with allocated budget $B'_p$}
    \State Initialize $z \gets z_{\text{init}}$, $\mathcal{S}'_p \gets \emptyset$
    
    \While{$|\mathcal{S}'_p| < B'_p$} \Comment{Continue until budget is met}
        \State \textbf{Step 2.1: Adaptive Thresholding}
        \State Compute mean ($\mu$) and std.\ dev.\ ($\sigma$) of entropy ($\mathbf{H}$) resp.\ similarity ($\mathbf{S}$) of the sample pools:
        \State \hspace{1em} $\mu_\text{TP}, \sigma_\text{TP} \gets$ Mean, std.\ dev.\ of $\mathbf{H}$ in $\text{TP}_{\text{pool}}$
        \State \hspace{1em} $\mu_\text{FN}, \sigma_\text{FN} \gets$ Mean, std.\ dev.\ of $\mathbf{H}$ in $\text{FN}_{\text{pool}}$
        \State \hspace{1em} $\mu_\text{FP}, \sigma_\text{FP} \gets$ Mean, std.\ dev.\ of $\mathbf{S}$ in $\text{FP}_{\text{pool}}$
        
        \State Compute sampling thresholds:
        \State $h_\text{TP} \gets \mu_\text{TP} + z \sigma_\text{TP}, \quad h_\text{FN} \gets \mu_\text{FN} + z \sigma_\text{FN}$
        \State $t_\text{FN} \gets \mu_\text{FN} - z \sigma_\text{FN}, \quad t_\text{FP} \gets \mu_\text{FP} - z \sigma_\text{FP}$
        
        \State \textbf{Step 2.2: Sampling Criteria} 
        \State Select samples ($\mathbf{s}$) with high-entropy TP, high-entropy FN, low-entropy FN, and low-similarity FP
        \State $\mathcal{S}'_p \gets \mathcal{S}'_p \cup \{ \mathbf{s} \in \text{TP}_{\text{pool}} \mid \mathbf{H}(\mathbf{s}) > h_\text{TP} \}$
        \State $\mathcal{S}'_p \gets \mathcal{S}'_p \cup \{ \mathbf{s} \in \text{FN}_{\text{pool}} \mid \mathbf{H}(\mathbf{s}) > h_\text{FN} \}$
        \State $\mathcal{S}'_p \gets \mathcal{S}'_p \cup \{ \mathbf{s} \in \text{FN}_{\text{pool}} \mid \mathbf{H}(\mathbf{s}) < t_\text{FN} \}$
        \State $\mathcal{S}'_p \gets \mathcal{S}'_p \cup \{ \mathbf{s} \in \text{FP}_{\text{pool}} \mid \mathbf{S}(\mathbf{s}) < t_\text{FP} \}$

        \State \textbf{Step 2.3: Iterative Threshold Refinement}
        \If{$|\mathcal{S}'_p| < B'_p$} 
            \State $z \gets z - 0.1$ \Comment{adjust threshold if insufficient samples}
        \EndIf
    \EndWhile
    
    \State \textbf{Step 3: Updating the Training Set}
    \State $\mathcal{S}_{\text{train}} \gets \mathcal{S}_{\text{train}} \cup \mathcal{S}'_p$
    \State $\mathcal{S}_{\text{pool}} \gets \mathcal{S}_{\text{pool}} \setminus \mathcal{S}'_p$
\EndFor

\end{algorithmic}
\end{algorithm}

\textbf{Step 2.1: Adaptive thresholding.} Fixed thresholds may not generalize well across predicates due to varying entropy and similarity distributions. To address this, we estimate per-predicate sampling thresholds by fitting normal distributions to entropy $\mathbf{H}$ and similarity $\mathbf{S}$, computing means and standard deviations for TPs, FNs, and FPs (Lines 16–23). These statistics guide adaptive thresholding.

\textbf{Step 2.2: Sampling criteria.} Sample selection ensures that the most informative samples are chosen for fine-tuning.  
\emph{(i) True Positives (TPs):} We use entropy \(\mathbf{H}\) to measure uncertainty. High-entropy TPs (\(\mathbf{H}(\mathbf{s}) > h_\text{TP}\)) indicate cases where the model is uncertain despite correct predictions, suggesting further refinement (Line 26). Therefore, we sample these instances to enhance model confidence and robustness. Conversely, low-entropy TPs represent confidently correct predictions and do not require further sampling.    
\emph{(ii) False Negatives (FNs):} Both high-entropy (\(\mathbf{H}(\mathbf{s}) > h_\text{FN}\)) and low-entropy (\(\mathbf{H}(\mathbf{s}) < t_\text{FN}\)) FNs are valuable -- high-entropy FNs signal model confusion, meaning the model is uncertain about its incorrect predictions, necessitating further training. In contrast, low-entropy FNs expose overconfident misclassifications, where the model is incorrectly certain about a wrong prediction. Addressing both extremes enhances the model’s ability to distinguish correct and incorrect relationships (Lines 27–28).    
\emph{(iii) False Positives (FPs):} Instead of entropy, similarity \(\mathbf{S}\) is used, as FPs vary in their semantic closeness to the ground truth. High-similarity FPs (\(\mathbf{S}(\mathbf{s}) > t_\text{FP}\), \eg, \cref{fig:art}, instance \circled{A}{yellow}, where GT: ``girl with hair” and Pred: ``girl has hair”) retain meaningful semantics, making them less critical for correction. In contrast, low-similarity FPs (\(\mathbf{S}(\mathbf{s}) < t_\text{FP}\), \eg, \cref{fig:art}, instance \circled{B}{yellow}, where GT: ``bag on table” and Pred: ``bag under table”) indicate weak generalization and require refinement. Since similarity is undefined for FNs (due to missing predictions) and trivially 1 for TPs (as predictions match the ground truth), it is applied exclusively to FPs. We sample FPs below the similarity threshold to ensure that the model improves on the most misleading cases (Line 29).

\textbf{Step 2.3: Iterative threshold refinement.} If the selected samples fall short of the budget $B'_p$, the threshold $z$ is iteratively reduced to ensure sufficient yet controlled, per-predicate sampling (Lines 30–33).  

\textbf{Step 3: Updating the training set.} The selected samples $\mathcal{S}'_p$ are added to $\mathcal{S}_{\text{train}}$ and removed from $\mathcal{S}_{\text{pool}}$ to ensure non-redundant sampling in future iterations (Lines 35–37). 

Refer to Appendix \ref*{app:art_through_examples} for an intuitive overview of ART.

\section{Experiments}
\label{sec:exp}
We aim to answer the following questions: 
\textbf{(Q1)}~Does ART outperform its baselines on in-distribution samples with unseen predicates? \textbf{(Q2)}~How well does it handle out-of-distribution data? \textbf{(Q3)} What happens when the complexity of out-of-distribution data increases? \textbf{(Q4)} How effective is the adaptive relation tuning framework? 
\textbf{(Q5)} How robust and beneficial is ART in real-world settings involving downstream tasks and noisy object detections?

\subsection{Experimental setup}

\textbf{Datasets.} We train on Visual Genome (VG) \cite{krishna2017visual} and test on GQA \cite{hudson2019gqa} and Open Images (OI) v4, v6 \cite{OpenImages} to evaluate model generalization across datasets with increasing complexity and out-of-distribution scenarios. 
Since VG is a subset of GQA with overlapping categories, it is easier than OI-v4 and v6, which are entirely distinct from VG and GQA.
See Appendix \ref*{sec:additional_exp_details} for data splits and further details.

\inparagraph{Evaluation protocol and metrics.} %As our aim is to enhance generalized predicate prediction, we assess the models' performance on the predicate classification task, using ground-truth bounding boxes.
As our aim is to improve predicate prediction in a generalizable manner, we evaluate model performance on the predicate classification task using ground-truth/predicted bounding boxes and objects, with a focus on generalization across data distributions, objects, and predicate categories. Following previous works \cite{zellers2018neural, sudhakaran2023vision}, we report Recall@k (R@k) and mean Recall (mR@k). \emph{Mean recall (mR@k) is particularly important} as it reflects a model’s ability to perform across the full predicate distribution.
To further assess generalization, we propose generalized Recall (gR@k) and mean generalized Recall (mgR@k), which treat false positives with high semantic similarity (measured by $\mathbf{S}$ from \cref{sec:art}) as true positives. See Appendix \ref*{sec:additional_exp_details} for details.

\inparagraph{Implementation details.} We leverage BLIP-2~\cite{li2023blip} models fine-tuned for the captioning task with strong generalization capabilities, specifically two InstructBLIP \cite{dai2024instructblip} variants. Both share the ViT/14 \cite{fang2023eva} image encoder but differ in language models:  One integrates a Vicuna-7B~\cite{zheng2023judging} LLM, instruction-tuned from LLaMA~\cite{touvron2023llama}, while the other uses FlanT5~\cite{chung2024scaling}, based on Transformer T5~\cite{raffel2020exploring}. 
We fine-tune both variants with our adaptive instruction tuning.  
The models reached the best results with about $12\%$ (see Appendix \ref*{sec:additional_exp_details}) of the training data samples, with a sampling budget of $2\%$ per adaptive learning loop. 

\subsection{Experimental results}
\label{subsec:exp_results}

Using the above experimental setup, we are able to address the questions \textbf{(Q1--5)}.
We establish fair comparisons by using several baselines, including naive relation-tuned VLMs (Naive-RT) with random and balanced-random sampling setups. Naive-RT (random) selects training samples through basic random sampling, while Naive-RT (balanced random) ensures even sampling across predicate categories, with random selection within each category. Additionally, we evaluate mainstream VRD models (Motifs~\cite{zellers2018neural}, VTransE~\cite{zhang2017visual}) and recent models with balanced mR (PENET~\cite{zheng2023prototype}, VETO~\cite{sudhakaran2023vision}). Higher mR reflects diverse relational learning, avoiding overfitting to dominant predicates. To assess generalization, all models are trained on VG and evaluated on GQA and OI.

\begin{figure}[tb]
    \centering
    \includegraphics[width=0.99\linewidth]{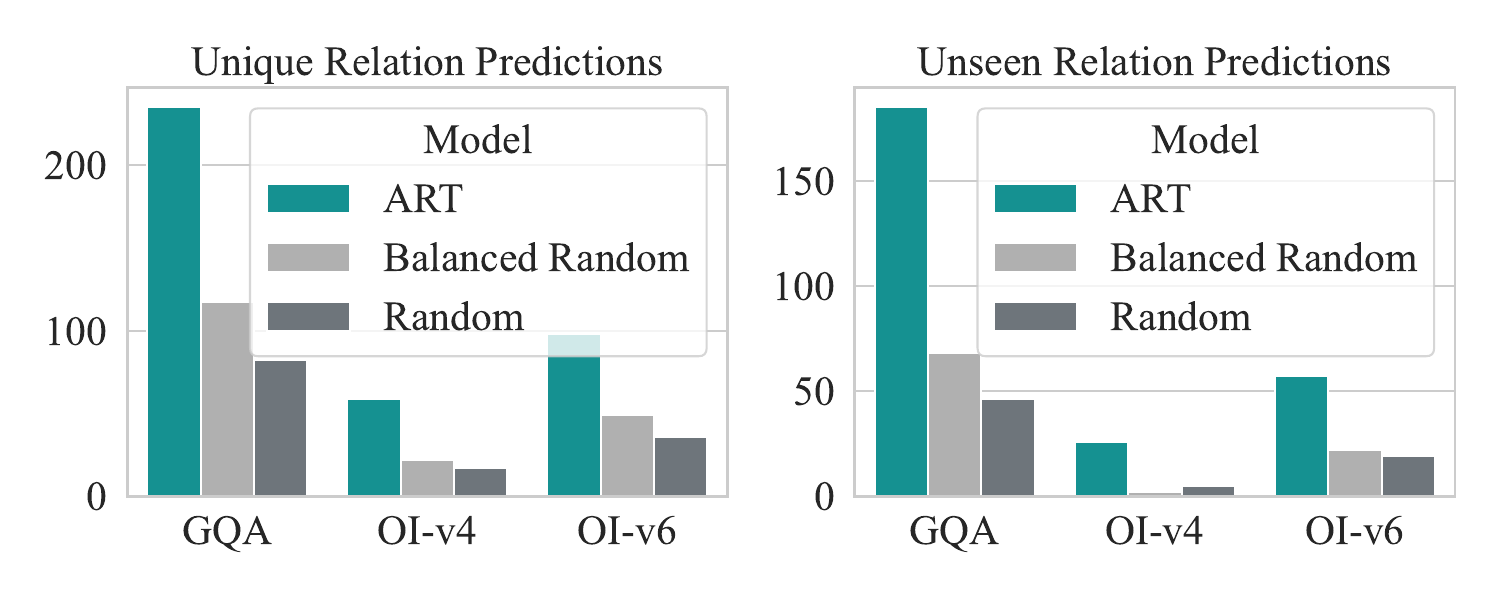}
    \vspace{-1em}
    \caption{\textbf{Comparison of unique relation predictions} \emph{(left)} \textbf{and unseen relation predictions} \emph{(right)} for the ART (Vicuna-7B) model across different datasets.}
    \vspace{-0.5em}
    \label{fig:vicuna_predictions}
\end{figure}

\inparagraph{(Q1) In-distribution samples with unseen predicates.} We analyze the methods in a simpler generalization setting using GQA, which shares images with VG (in-distribution) but has more object classes and predicates. Like VG, GQA is dominated by a few head predicates. Consequently, models that generalize well should achieve higher mR and gmR, indicating diverse relational learning rather than overfitting to head categories, which results in high R but low mR.  As seen in \cref{tab:results}, Vicuna-7B+ARToutperforms the strongest baseline, Vicuna-7B+Naive-RT (balanced random), with more than a 40\% and 20\% relative improvement in mR and gmR, respectively. Similar gains are observed for FlanT5. In contrast, mainstream models and Naive-RT (random) achieve high R but lower mR, revealing their bias toward frequent predicates. 
Additional evaluations (\cref{fig:vicuna_predictions})  confirm ART’s superior relational diversity, with more unique and unseen predicates than baselines.

\begin{table*}[]
\caption{\textbf{Evaluation of VG-trained models on GQA, Open Images v4 (OI-v4), and Open Images v6 (OI-v6),} showing Recall (R), mean Recall (mR), and their generalized forms (gR, mgR, see text) — all in \%, higher values indicate better performance.   The superscripts `†' and `*' denote methods that use TDE and re-weighting strategies, respectively. Double citations refer to the original model and the used TDE-debiased version. The superscript `‡' denotes models using VLMs with the same image encoder (ViT-g/14) but different LLMs: \textcolor{cyan}{FlanT5}~\cite{chung2024scaling} and \textcolor{gray}{Vicuna}~\cite{zheng2023judging}. The best results for each VLM variant are highlighted in \textbf{bold}.}
\centering
\footnotesize
\vspace{-0.5em}
\begin{tabularx}{\linewidth}{>{\hspace{-\tabcolsep}\raggedright\columncolor{white}[\tabcolsep][\tabcolsep]}Xlcccc}
\toprule
%Dataset     & \multicolumn{4}{c|}{GQA}                                                                                           \\ \hline
Model & Dataset \qquad\qquad & R@k: 20/50 & mR@k: 20/50 & gR@k: 20/50 & gmR@k: 20/50   \\
\midrule %%%%%%%%%%%%%%%%%%%%%%%%%%%%%%%%%%%%%%%
Motifs\textsuperscript{†}~\cite{tang2020unbiased,zellers2018neural}
& & 41.7 / 41.7 & 12.7 / 12.8 & 55.7 / 55.8 & 23.5 / 23.6 \\ 
VTransE\textsuperscript{†}~\cite{tang2020unbiased,zhang2017visual}     
& &  37.9 / 38.0      &  9.8 / 9.9          &  51.3 / 51.3        &  18.7 / 18.8            \\ 
PENET\textsuperscript{*}~\cite{zheng2023prototype}       
& & 45.8 / 45.8 & 10.2 / 10.3 & 62.7 / 62.7 & 21.7 / 21.8 \\ 
VETO\textsuperscript{*}~\cite{sudhakaran2023vision}        
& & 47.7 / 47.8 & 10.9 / 10.9 & 63.4 / 63.5 & 23.2 / 23.3 \\ 

\rowcolor{gray!20}
FlanT5-XL\textsuperscript{‡} + Naive-RT (random) 
&  & \textbf{51.4} / \textbf{53.4} & 8.9 / 9.5 & \textbf{65.8} / \textbf{69.5} & 20.0 / 22.6 \\
\rowcolor{gray!20}
FlanT5-XL\textsuperscript{‡} + Naive-RT (balanced random) 
&  &  36.1 / 37.7 & 13.6 / 14.7 & 53.7 / 56.8 & 25.6 / 29.2 \\
\rowcolor{gray!20}
FlanT5-XL\textsuperscript{‡} + \textbf{ART (ours)} 
& & 30.3 / 31.5 & \textbf{15.5} / \textbf{16.1} & 61.1 / 62.0 & \textbf{30.8} / \textbf{32.2} \\

\rowcolor{cyan!20}
Vicuna-7B\textsuperscript{‡} + Naive-RT (random) 
&  & \textbf{59.0} / \textbf{61.3} & 8.3 / 9.1 & \textbf{70.0} / \textbf{73.9} & 18.8 / 21.7 \\
\rowcolor{cyan!20}
Vicuna-7B\textsuperscript{‡} + Naive-RT (balanced random) 
&  & 31.9 / 33.0 & 13.2 / 14.1 & 55.1 / 58.3 & 26.0 / 29.7 \\
\rowcolor{cyan!20}
Vicuna-7B\textsuperscript{‡} + \textbf{ART (ours)} 
& \multirow{-10}{*}{GQA} & 40.1 / 40.4 & \textbf{18.9} / \textbf{19.4} & 61.5 / 62.2 & \textbf{33.2} / \textbf{34.7} \\

\midrule %%%%%%%%%%%%%%%%%%%%%%%%%%%%%%%%%%%%%%%
Motifs\textsuperscript{†}~\cite{tang2020unbiased,zellers2018neural}
& & 30.0 / 30.1 & 13.4 / 13.4 & 56.9 / 57.0 & 31.0 / 31.1 \\
VTransE\textsuperscript{†}~\cite{tang2020unbiased,zhang2017visual}     
& & 29.7 / 29.8 & 13.3 / 13.4 & 55.7 / 56.0  &  29.3 / 29.6           \\
PENET\textsuperscript{*}~\cite{zheng2023prototype}       
& & 2.1 / 2.2 & 0.6 / 0.7 & 18.1 / 18.2 & 0.6 / 0.7 \\
VETO\textsuperscript{*}~\cite{sudhakaran2023vision}        
& & 12.1 / 12.1 & 6.7 / 6.7 & 46.3 / 46.4 & 28.3 / 28.4 \\

\rowcolor{gray!20}
FlanT5-XL\textsuperscript{‡} + Naive-RT (random) 
&  & 30.6 / 31.8 & 14.5 / 14.9 & 70.0 / 72.5 & 40.5 / 40.9 \\
\rowcolor{gray!20}
FlanT5-XL\textsuperscript{‡} + Naive-RT (balanced random) 
&  &  40.1 / 42.2 & 15.0 / 15.6 & 69.7 / 72.2 & 40.0 / 40.9 \\
\rowcolor{gray!20}
FlanT5-XL\textsuperscript{‡} + \textbf{ART (ours)} 
& & \textbf{54.0} / \textbf{54.7} & \textbf{17.1} / \textbf{17.3} & \textbf{81.0} / \textbf{82.2} & \textbf{44.9} / \textbf{45.4} \\

\rowcolor{cyan!20}
Vicuna-7B\textsuperscript{‡} + Naive-RT (random) 
&  & 43.5 / 44.6 & 12.7 / 13.1 & 73.9 / 76.0 & 37.8 / 38.6 \\
\rowcolor{cyan!20}
Vicuna-7B\textsuperscript{‡} + Naive-RT (balanced random) 
&  & 42.5 / 44.7 & 11.8 / 12.4 & 72.4 / 74.9 & 35.5 / 36.5 \\
\rowcolor{cyan!20}
Vicuna-7B\textsuperscript{‡} + \textbf{ART (ours)} 
& \multirow{-10}{*}{OI-v4} & \textbf{46.6} / \textbf{47.5} & \textbf{26.0} / \textbf{26.2} & \textbf{79.0} / \textbf{80.4} & \textbf{54.3} / \textbf{54.5} \\

\midrule %%%%%%%%%%%%%%%%%%%%%%%%%%%%%%%%%%%%%%%
Motifs\textsuperscript{†}~\cite{tang2020unbiased,zellers2018neural}
& & 6.3 / 6.3 & 4.0 / 4.1 & 54.8 / 54.9 & 13.9 / 13.9 \\ 
VTransE\textsuperscript{†}~\cite{tang2020unbiased,zhang2017visual}    
&  & 4.5 / 4.6        &   2.9 / 3.0           &   27.4 / 27.6          &  9.2 / 9.2           \\ 
PENET\textsuperscript{*}~\cite{zheng2023prototype}       
& & 1.3 / 1.4 & 0.1 / 0.1 & 19.9 / 19.9 & 5.9 / 6.0 \\ 
VETO\textsuperscript{*}~\cite{sudhakaran2023vision}        
& & 3.4 / 3.4 & 2.5 / 2.5 &50.0 / 50.0 & 12.2 / 12.2 \\

\rowcolor{gray!20}
FlanT5-XL\textsuperscript{‡} + Naive-RT (random) 
&  & 7.2 / 7.4 & 5.5 / 5.6 & 54.5 / 55.2 & 21.7 / 21.9 \\
\rowcolor{gray!20}
FlanT5-XL\textsuperscript{‡} + Naive-RT (balanced random) 
&  &  19.6 / 19.9 & 7.0 / 7.1 & 56.0 / 56.8 & 22.0 / 22.3 \\
\rowcolor{gray!20}
FlanT5-XL\textsuperscript{‡} + \textbf{ART (ours)} 
& & \textbf{21.0} / \textbf{21.2} & \textbf{10.4} / \textbf{10.5} & \textbf{57.4} / \textbf{57.7} & \textbf{25.5} / \textbf{25.7} \\

\rowcolor{cyan!20}
Vicuna-7B\textsuperscript{‡} + Naive-RT (random) 
&  & 9.7 / 9.8 & 3.8 / 3.9 & 56.3 / 57.5 & 15.4 / 16.7 \\
\rowcolor{cyan!20}
Vicuna-7B\textsuperscript{‡} + Naive-RT (balanced random) 
&  & 23.4 / 25.0 & 8.5 / 8.7 & 55.0 / 55.8 & 22.1 / 22.5 \\
\rowcolor{cyan!20}
Vicuna-7B\textsuperscript{‡} + \textbf{ART (ours)} 
& \multirow{-10}{*}{OI-v6} & \textbf{27.3} / \textbf{27.4} & \textbf{9.5} / \textbf{9.6} & \textbf{63.2} / \textbf{63.4} & \textbf{25.6} / \textbf{25.8} \\

\bottomrule
\end{tabularx}
\label{tab:results}
\vspace{-0.5em}
\end{table*}

\begin{table}[tb]
\caption{\textbf{Instruction set analysis of ART (Vicuna-7B) on VG.} \emph{Rel} indicates relations, and \emph{Neg} denotes negative samples. Removing components lowers Recall (R@k) and mean Recall (mR@k). High-level relations slightly boost both metrics, while counter-negatives notably improve mR@k. Best results are in \textbf{bold}. The proposed final instruction set is highlighted in \textcolor{cyan}{cyan}.}
\vspace{-0.5em}
\centering
\footnotesize
\setlength{\tabcolsep}{3pt} % Reduce column padding
\renewcommand{\arraystretch}{0.95} % Reduce row height
\begin{tabularx}{\linewidth}{@{}>{\centering\arraybackslash}p{1.2cm} >{\centering\arraybackslash}p{1.1cm} >{\centering\arraybackslash}p{1.1cm} >{\centering\arraybackslash}p{1.5cm} >{\centering\arraybackslash}p{1.2cm} >{\centering\arraybackslash}p{1.2cm}@{}}
\toprule
High-level \emph{Rel} & Random \emph{Neg} & Counter \emph{Neg} & Per-sample \emph{Neg} & R@k (20/50) & mR@k (20/50) \\
\midrule
\n & \n & \n & -- & 37.5/38.0 & 37.5/38.0\\ 
\y & \n & \n & -- & 38.6/39.1 & 39.4/39.9\\
\y & \y & \n & 1 & \textbf{41.2}/\textbf{41.8} & 40.9/41.3\\ 
\y & \y & \n & 2 & 27.9/28.0 & 41.3/41.7\\ 
%\rowcolor{cyan!20} 
\y\cellcolor{cyan!20} & \n\cellcolor{cyan!20} & \y\cellcolor{cyan!20} & 1\cellcolor{cyan!20} & 41.1/41.4\cellcolor{cyan!20} & \textbf{46.4}/\textbf{47.7}\cellcolor{cyan!20}\\ 
\y & \n & \y & 2 & 37.0/37.3 & 40.3/41.4\\
\bottomrule
\end{tabularx}
\label{tab:instruction_set_ablation}
\vspace{-1em}
\end{table}

\inparagraph{(Q2) Out-of-distribution samples with unseen predicates.}In this setting, we assess a more challenging generalization scenario using the OI-v4 dataset. Since OI contains distinct image samples from VG, it is considered out-of-distribution data and includes unseen object and predicate categories. As shown in \cref{tab:results}, ART consistently outperforms its baselines across all metrics for both the Vicuna and FlanT5 model variants. 
Unique and unseen predictions on OI-v4 (\cref{fig:vicuna_predictions}) further confirm ART's ability to predict diverse relations in out-of-distribution settings.

\inparagraph{(Q3) Out-of-distribution samples with increased complexity.} In this scenario, we further evaluate the out-of-distribution generalization of ART on OI-v6. As shown in \cref{tab:results}, mainstream models drop significantly across all metrics, while ART remains robust, consistently improving mR and gmR over its baselines. Notably, FlanT5+ART achieves a 50\% mR improvement over the second-best Naive-RT (balanced random), reinforcing ART's effectiveness in complex out-of-distribution settings.

\inparagraph{(Q4) Effectiveness of adaptive relation tuning (ART) framework.} We conduct three analyses to evaluate the effectiveness of the ART framework.

\textbf{Instruction set analysis.} We examine the impact of our proposed instruction set from \cref{sec:art:data}. As shown in \cref{tab:instruction_set_ablation}, incorporating high-level relation types (2\textsuperscript{nd} row) in the prompt improves both R and mR.  Adding a single random negative response per positive response (3\textsuperscript{rd} row) further enhances these metrics. However, increasing the number of random negatives leads to a notable drop in R.  Replacing random negatives with a single counter-negative response yields the highest mR; adding more counter-negatives reduces both R and mR. Thus, we use one counter-negative per sample in our final instruction set.

 \textbf{Sampling component analysis.} We compare our full ART model with all sampling components in \cref{tab:ablation_art}, (\ie $h_{\text{FN}}$, $t_{\text{FP}}$, $t_{\text{FN}}$, $h_{\text{TP}}$, last row) against its variants. When comparing a model trained on the full VG train instructions (1\textsuperscript{st} row) to a model trained on a reduced random 12\% sampling budget (2\textsuperscript{nd} row), the model trained on a random subset outperforms the fully trained one.  This indicates that tuning a VLM on more data can lead to suboptimal results if the data distribution is biased. Comparing  random to balanced random sampling reveals that the balanced random model better learns various relation concepts, resulting in improved mR. Introducing our ART strategy results in further improved mR. The ART: \emph{entropy} model that picks the highest entropy samples from the entire sampling pool is less efficient than the variants that diligently pick based on the uncertainty or dissimilarity ($\mathbf{H}$ or $\mathbf{S}$) and the head or tail region to sample from the prediction types (TP, FP, or FN).

Additionally, a detailed analysis of adaptive \vs fixed thresholding is provided in Appendix \ref*{app:thresholding}.

\begin{table}[]
\caption{\textbf{Ablation study of ART on VG.} Adaptive relation tuned model variants. \emph{entropy}: only entropy-based uncertainty is used, $h_{\text{TP}}$: entropy-based head samples from TP, $h_{\text{FN}}$: entropy-based head samples from FN, $t_{\text{FN}}$: entropy-based tail samples from FN, $t_{\text{FP}}$: similarity score-based tail samples from FP. TP: True Positive, FN: False Negative, FP: False Positive. Except the 1\textsuperscript{st} row that uses 100\% train data, all other models use 12\% of train data. The proposed final ART model is highlighted in \textcolor{cyan}{cyan}. }
\centering
\footnotesize
\vspace{-0.5em}
\begin{tabularx}{\linewidth}{>{\hspace{-\tabcolsep}\raggedright\columncolor{white}[\tabcolsep][\tabcolsep]}Xcc}
\toprule
Sampling strategy & R@k: 20/50   & mR@k: 20/50  \\
\midrule
full train data &  59.9 / 60.2 & 14.1 / 14.7 \\
random & \textbf{63.2} / \textbf{63.4} & 19.6 / 19.8 \\
balanced random & 36.4 / 36.6 & 42.4 / 43.5 \\
ART: \emph{entropy} & 34.7 / 34.9 & 44.6 / 44.8 \\
ART: $h_{\text{FN}}$, $t_{\text{FP}}$ &37.5 / 37.6 & 45.5 / 45.6 \\
ART: $h_{\text{FN}}$, $t_{\text{FP}}$, $t_{\text{FN}}$ &38.3 / 38.5 & 45.2 / 45.4 \\
\rowcolor{cyan!20} 
ART: $h_{\text{FN}}$, $t_{\text{FP}}$, $t_{\text{FN}}$, $h_{\text{TP}}$ & 41.1 / 41.4 & \textbf{46.4} / \textbf{47.7} \\ 
\bottomrule
\end{tabularx}
\label{tab:ablation_art}
\vspace{-0.5em}
\end{table}

%\inparagraph{(Q5) ART for downstream image segmentation.}
\inparagraph{(Q5) Robustness across tasks and inputs.}
We evaluate ART’s effectiveness in a downstream application, specifically a \emph{deictic}\footnote{A deictic representation refers to an object depending on the overall context, \eg, ``An object on the table and next to a cup.''} segmentation task~\cite{shindo2024deisam} using ART.
Given visual input and a complex textual prompt, the task is to produce a segmentation mask for the object specified by the prompt.
We evaluate ART on a deictic segmentation task~\cite{shindo2024deisam} using 10k image–prompt pairs with expressive relations absent from training data (\eg, \emph{floating on}, \emph{attached to}). We compare: (1) LISA~\cite{lai2023lisa}, a state-of-the-art VLM-based reasoning model; (2) DeiSAM, a neuro-symbolic segmentation model using ground-truth scene graphs (DeiSAM + GT); and using ART-generated graphs (DeiSAM + ART). Results with and without ground-truth boxes are reported. As shown in \cref{tab:mAP_results}, ART outperforms all baselines, demonstrating strong generalization to novel relations, especially valuable in dynamic environments.

\begin{table}[t]
    \centering
    \caption{\textbf{ART enhances segmentation qualities.} Mean Average Precision (mAP) comparison of DeiSAM+ART, DeiSAM+GT, and LISA on VG and GQA datasets.}
    \footnotesize
    \vspace{-0.5em}
    \begin{tabularx}{\linewidth}{@{}XS[table-format=2.2]S[table-format=2.2]@{}}
        \toprule
        mAP (in \%, $\uparrow$) & VG  & GQA \\        \midrule
        LISA~\cite{lai2023lisa}          & 7.87 & 18.92 \\
        DeiSAM + GT (w/o bbox)   & 4.77 & 11.84 \\
        DeiSAM + ART (w/o bbox)           & \bfseries 25.07 & \bfseries 26.62 \\ \midrule
        DeiSAM + GT (with bbox)   & 35.04 & 43.84 \\
        DeiSAM + ART (with bbox)           & \bfseries 99.98 & \bfseries 97.96 \\
        \bottomrule
    \end{tabularx}
    \label{tab:mAP_results}
    \vspace{-0.5em}
\end{table}

To further assess robustness, we evaluate ART using noisy object detections from real-world detectors. Specifically, we test with boxes from Detectron2~\cite{wu2019detectron2} and the zero-shot RAM model~\cite{zhang2024recognize}. As shown in \cref{tab:detected-results}, ART maintains strong performance across both detectors. Remarkably, ART significantly outperforms its baselines even when using RAM’s open-world detections, confirming its adaptability and resilience to detection noise.

\begin{table}[h]
\centering
\footnotesize
\caption{\textbf{Performance on detected bounding boxes} from  \textcolor{gray}{Detectron2}~\cite{wu2019detectron2} and \textcolor{cyan}{RAM}~\cite{zhang2024recognize}. R: Random sampling, B: Balanced Random sampling, and ART (ours).}
\vspace{-1em}
\begin{tabularx}{\linewidth}{X X CCCC}
\toprule
\textbf{Model} & \textbf{Dataset} & \textbf{R@20} & \textbf{mR@20} & \textbf{gR@20} & \textbf{gmR@20} \\
\midrule

\rowcolor{gray!20}
R         &  &   42.6     &  13.0      &  64.0      & 24.5   \\
\rowcolor{gray!20}
BR       & OI-v4      & 35.3       &   11.1     & 51.1       &   22.6     \\
\rowcolor{gray!20}
ART        &    &  \textbf{53.7}    &  \textbf{16.3}       &  \textbf{69.5}       &  \textbf{26.8}              \\
\midrule
\rowcolor{cyan!20}
R        &  & 19.1  & 6.7       & 55.0       &  22.3      \\
\rowcolor{cyan!20}
BR       &      OI-v4        & 29.2  & 7.3       & 59.4       & 26.5       \\
\rowcolor{cyan!20}
ART            &              & \textbf{40.6}  & \textbf{24.0}       & \textbf{69.1}       & \textbf{42.6}       \\

\bottomrule
\end{tabularx}
\label{tab:detected-results}
%\vspace{-1em}
\end{table}

\section{Conclusion}
In this work, we presented ART, a novel framework for the relation classification task of VRD that leverages vision-language models through instruction tuning by converting benchmark VRD datasets into an instruction-tuning format. Our model focuses on the most relevant instances by incorporating adaptive sampling and fine-tuning techniques, thus enhancing its robustness and generalization capabilities. ART consistently outperforms baseline models, improving the handling of both diverse and unseen object classes and relations. This underscores the importance of innovative VRD approaches, especially for real-world applications with diverse data distributions. 
%ART's success in inferring relations in complex visual scenes and generalizing to novel instances marks a substantial advancement in the field. 
Future research can enhance ART by refining instruction tuning strategies and extending it to other vision-language tasks, potentially unlocking new applications in visual understanding.

{\small \inparagraph{Acknowledgments.} This work was funded by the Hessian Ministry of Science and the Arts (HMWK) through the
projects “The Third Wave of Artificial Intelligence -- 3AI” and hessian.AI. 
The work was further supported by the Deutsche Forschungsgemeinschaft (German Research Foundation, DFG) under Germany’s Excellence Strategy (EXC 3057/1 “Reasonable Artificial Intelligence”, Project No.\ 533677015). 
SR acknowledges support by the European Research Council (ERC) under the European Union’s Horizon 2020 research and innovation programme (grant agreement No.\ 866008).
}

{
    \small
    \bibliographystyle{ieeenat_fullname}
    \bibliography{main}
}

\clearpage
\clearpage
\setcounter{section}{0}
\renewcommand\thesection{\Alph{section}}
\setcounter{page}{1}
\pagenumbering{roman}
\twocolumn[{
  \renewcommand\twocolumn[1][]{#1}
  \begin{center}
    \Large \textbf{ART: Adaptive Relation Tuning for Generalized Relation Prediction}\\[0.5em]
    \large \textbf{Supplementary Material}\\[1.2em]
    
    \normalsize
    {\large
      Gopika Sudhakaran$^{1,2}$ \quad
      Hikaru Shindo$^1$ \quad
      Patrick Schramowski$^{1,3}$ \quad
      Simone Schaub-Meyer$^{1,2}$ \quad
      Kristian Kersting$^{1,2,3}$ \quad
      Stefan Roth$^{1,2}$\\[0.5em]
    }
    $^1$Department of Computer Science, TU Darmstadt, Germany\\
    $^2$Hessian Center for AI (hessian.AI) \quad
    $^3$German Research Center for AI (DFKI)\\[1em]
  \end{center}
}]

\section{Overview}
We provide supplementary experimental details, followed by an in-depth explanation of the balanced sampling algorithm used for initial sampling.  We then include a dedicated section — \textit{Understanding ART through examples} — which offers an intuitive walkthrough of ART’s core sampling strategy using illustrative cases. Next, we analyze the relation predictions based on their diversity and whether they are unseen (\ie, not contained in the training annotations). We then discuss the computational cost of ART and its baselines, analyze the trade-off between data usage and performance, and clarify the behavior of ART on certain recall metrics. Finally, we conclude with a qualitative comparison between ART and its baselines. 

\section{Additional experimental details}
\label{sec:additional_exp_details}
\begin{figure}[b]%
\vspace{-0.5em}

\centering

%\hspace{-2em}
\includegraphics[width=0.99\linewidth]{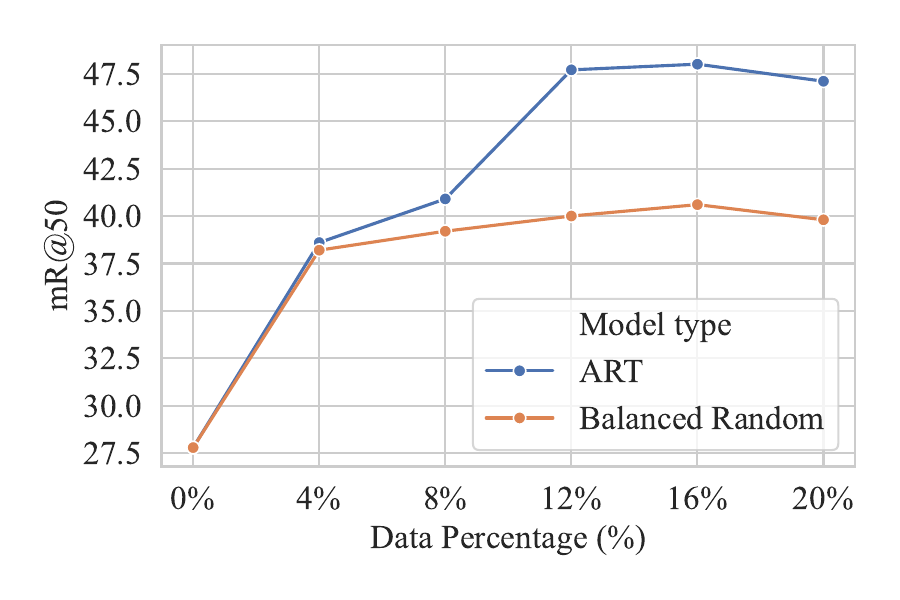}
%\decoRule
\vspace{-1.25em}
\caption{\textbf{Training data subsampling analysis.} We plot mR@50 for ART and Naive-RT (balanced random) as a function of the training data percentage used (y-axis) on Vicuna model variants.}
\label{fig:data_usage}
\end{figure}

\paragraph{Training details.} In addition to the hyperparameters outlined in \cref{sec:exp} of the main paper, we set the initial $z$-score threshold to 1.96, which corresponds to 95\% of the data. This threshold was chosen because a $z$-score of 1.96 is more sensitive to potential outliers, making it useful for capturing more subtle deviations from the norm. For training, we use an initial learning rate of $1e-3$ and a linear warmup for 3000 steps. We optimize with Adam ($\beta_{1}=0.9$, $\beta_{2}=0.999$) and apply a weight decay of 0.05.  Additionally, we chose 12\% of training data for instruction tuning as mR@k saturates near this point as analyzed in \cref{fig:data_usage}. We use the LAVIS library~\cite{li2022lavis} for implementation, training, and evaluation. 
The models are trained using four Nvidia A100 (40Gb) GPUs within two days.

\inparagraph{Dataset details.}
We adopt the VG150 split for Visual Genome (VG) \cite{krishna2017visual}, which includes 150 object classes and 50 predicates, aligned with established baselines \cite{zellers2018neural, zhang2017visual, tang2020unbiased, zheng2023prototype, sudhakaran2023vision}. In comparison, GQA \cite{hudson2019gqa} (GQA200 split) includes 100 predicates and 200 object classes. VG is a subset of GQA with overlapping categories. %—50 shared predicates and 150 shared object classes. 
Testing on the expanded set of GQA allows us to assess the model’s generalization to new predicates and object categories, a more rigorous test of robustness than the reverse (training on GQA and testing on VG). Additionally, we test on Open Images (OI) \cite{OpenImages}, where OI-v4 includes 9 predicates and 57 object classes, and OI-v6 expands to 31 predicates and 601 object classes. Since OI’s data distribution is entirely distinct from VG and GQA, it serves as a fully out-of-distribution benchmark, presenting increased complexity and enabling us to comprehensively evaluate model adaptability and robustness to unseen categories and relationships.

\paragraph{On semantic similarity for evaluation.}
We threshold the semantic similarity $\mathbf{S}$ at 95\% to ensure that only highly semantically similar predictions are counted. For example, in \cref{fig:art}, FPs such as \circled{A}{yellow} that are semantically similar to the ground truth are counted as TPs. The similarity is computed over subject–predicate–object triplets with only the predicate varying. Even small differences (\eg, “bag on table” \vs “bag under table”) yield noticeable drops in similarity. The threshold 0.95 was selected based on qualitative analysis, which confirmed that high-similarity matches preserved meaningful semantics and did not introduce false positives. Notably, ART frequently predicts semantically rich alternatives (\eg, “girl petting dog” \vs GT: “girl interacts with dog”), which may be underrecognized by current metrics — pointing to potential improvements in future evaluation design.

\section{Balanced sampling}
\label{sec:balanced sampling}

As described in \cref{sec:art} of the main paper, the ART pipeline begins with a balanced sampling algorithm, described in \cref{alg:balanced_sampling}, to provide an unbiased and balanced understanding of the relations during the initial loop. This step ensures that the subsequent adaptive sampling loop is better guided to select informative samples rather than being influenced by the biases of the underlying data distribution. The balanced sampling distributes a fixed sampling budget across multiple predicates fairly. At first, the predicates are sorted in descending order of frequency, and their allocated budgets are set to zero. The algorithm then allocates a single sampling slot to a predicate with a non-zero frequency, decreases its frequency (\ie, availability), and reduces the remaining budget. If a predicate's availability is exhausted, the algorithm skips it and continues assigning slots to the remaining predicates in a round-robin manner.  This ensures that sampling focuses on predicates whose availability has not been exhausted while maintaining a balanced distribution as much as possible.

\begin{algorithm} [tb]
\caption{Balanced Sampling Module}
\label{alg:balanced_sampling}
\begin{algorithmic}[1]
\State \textbf{Input:} 
\State \hspace{4mm} $\mathcal{P} = \{ p_i \}_{i=1}^{N}$ \Comment{Set of $N$ predicate categories}
\State \hspace{4mm} $\mathcal{S}_{\text{train}} \gets \emptyset$ \Comment{Initial training set}
\State \hspace{4mm} $\mathcal{S}_{\text{pool}}$ \Comment{Remaining dataset (excluding $\mathcal{S}_{\text{train}}$)}
\State \hspace{4mm} $N_p \gets$ Available samples for each predicate $p_i$
\State \hspace{4mm} $B \gets$ Total sampling budget per iteration

\State \textbf{Output:} Updated training set $\mathcal{S}_{\text{train}}$ and pool set $\mathcal{S}_{\text{pool}}$

\State \textbf{Initialization:}
\State \hspace{4mm} $B'_p \gets 0, \quad \forall p \in \mathcal{P}$ \Comment{Initialize per-predicate budget}
\State \hspace{4mm} $i \gets 1$ 

\While{$B > 0$}
    \If{$N_{p_i} > 0$} 
        \State $B'_{p_i} \gets B'_{p_i} + 1$ \Comment{Allocate one sample to predicate $p_i$}
        \State $N_{p_i} \gets N_{p_i} - 1$ \Comment{Decrement available samples}
        \State $B \gets B - 1$ \Comment{Reduce remaining budget}
    \EndIf
    \State $i \gets (i + 1) \mod N$ \Comment{Move to the next predicate}
\EndWhile

\State \textbf{Assign samples to training set:}
\For{each predicate $p_i \in \mathcal{P}$}
    \State $\mathcal{S}_{\text{train}} \gets \mathcal{S}_{\text{train}} \cup$ \{Randomly selected $B'_{p_i}$ samples from $\mathcal{S}_{\text{pool}}\}$
    \State $\mathcal{S}_{\text{pool}} \gets \mathcal{S}_{\text{pool}} \setminus \mathcal{S}_{\text{train}}$
\EndFor

\end{algorithmic}
\end{algorithm}

\section{Understanding ART through examples}
\label{app:art_through_examples}
To help understand the inner workings of Adaptive Relation Tuning (ART), we illustrate the core sampling choices that guide learning. ART’s goal is to adapt vision-language models for robust and generalizable visual relation detection. It does this by selecting training instances that are not only informative but also help the model learn from its weaknesses.

We categorize predictions into three groups — True Positives (TP), False Negatives (FN), and False Positives (FP) — and strategically sample from each using a combination of entropy (model uncertainty) and semantic similarity (to ground truth). Below, we explain the reasoning behind each sampling choice with concrete examples:

\subsection{High-Entropy True Positives (TPs): Improve uncertain correct predictions}
These are predictions where the model gets the relation right, but shows uncertainty (high entropy) in doing so. Including them in training reinforces correct behavior and improves model confidence.

\textit{Example:} The model correctly predicts ``boy riding bike'' but assigns nearly equal probability to ``boy on bike''. This shows uncertainty despite being correct. Sampling this TP helps the model reinforce the right prediction with more certainty.
\subsection{Low- and High-Entropy False Negatives (FNs): Correct missed relations}

False Negatives occur when a relation exists in the ground truth, but the model says that no prominent relation exists.

\textit{Example:} If the ground truth is ``man holding umbrella'', the model may either hesitate (high entropy) or confidently predict ``no prominent relation exists'' (low entropy). Both cases are important — uncertain misses highlight confusion, while confident misses expose overfitting or bias. Sampling both types improves robustness.

\subsection{Low-Similarity False Positives (FPs): Penalize semantically incorrect predictions}

False positives are predicted relations that do not appear in the ground truth. However, not all FPs are equally harmful. Some are semantically close — or even more descriptive — and may still reflect a correct understanding of the scene. Others are misleading and indicate poor generalization.

\textit{Example:} Given the ground truth ``man in canoe'', predicting ``man under canoe'' is a low-similarity FP — it is spatially incorrect and misleading. On the other hand, predicting ``man paddling canoe'' is a high-similarity FP that, while not an exact match, is semantically rich and even more informative than the original label. ART distinguishes between such cases and focuses on refining the misleading ones.

\vspace{0.5em}
These sampling decisions are made adaptively per predicate using dynamically computed thresholds (based on per-predicate entropy and similarity distributions). This ensures flexible and targeted learning.

\section{Analysis of predicted relations}
\begin{figure}[tb]
    \centering
    \includegraphics[width=0.99\linewidth]{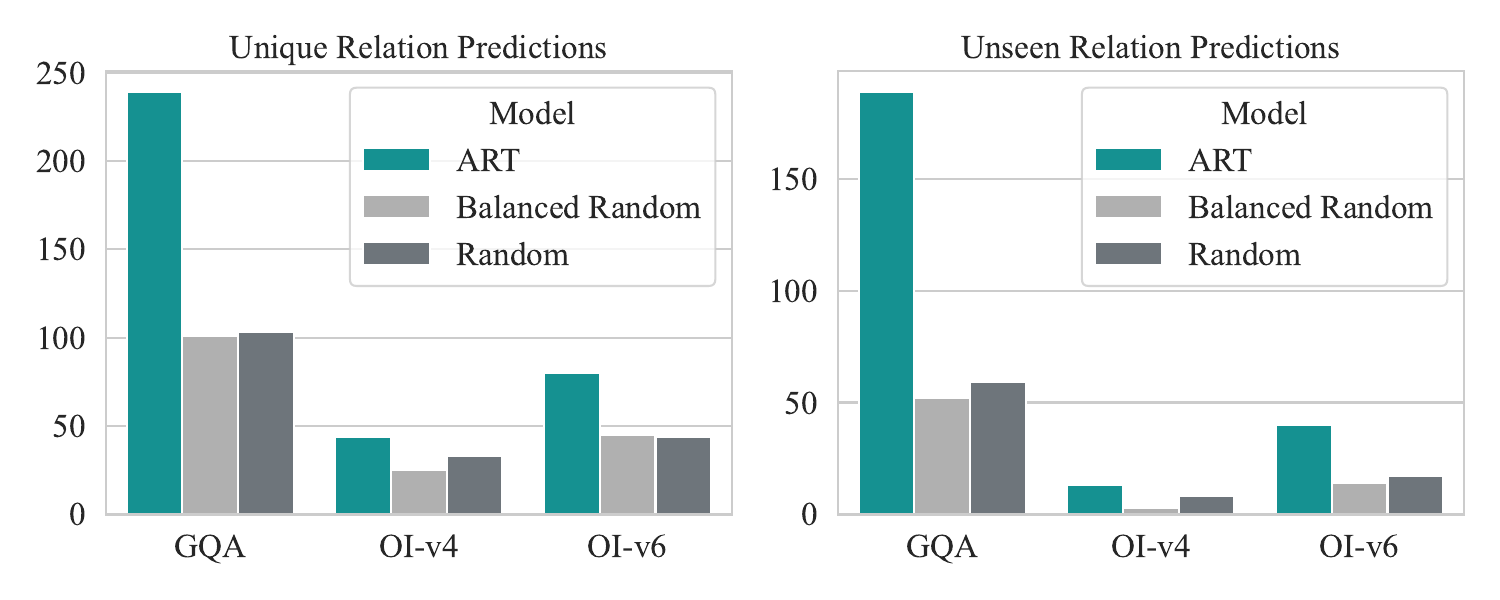}
    \vspace{-1em}
    \caption{\textbf{Comparison of unique relation predictions} \emph{(left)} \textbf{and unseen relation predictions} \emph{(right)} for the ART (FlanT5) model across different datasets.}
    \label{fig:flan_predictions}
    \vspace{-0.5em}
\end{figure}

To further evaluate the effectiveness of ART in predicting informative, diverse, and unseen relations, we compared its predictions against random and balanced random baseline methods for both Vicuna \cite{zheng2023judging} (see \cref{fig:vicuna_predictions}) and FlanT5 \cite{chung2024scaling} model variants. As discussed in \cref{sec:exp},  ART’s superiority in predicting diverse and unseen relations extends from Vicuna to FlanT5. \cref{fig:flan_predictions} (left) illustrates the total number of unique relations predicted by ART and its baselines. As shown, ART consistently predicts a greater variety of relations across all datasets. A similar pattern can be observed in \cref{fig:flan_predictions} (right), where ART predicts more relations unseen during training on VG.

Notably, GQA has the most test samples, followed by OI-v6 and OI-v4, leading to variations in total predictions. The larger GQA test set allows inference across broader scenarios, increasing the likelihood of predicting more diverse and unseen relations.

\begin{comment}
\begin{figure*}[ht]
    \centering
    \includegraphics[width=\linewidth]{figures/model_comparison.pdf}
    \caption{\textbf{Total number of unique relation predictions of Vicuna and FlanT5 model variants.} The plots depict the total number of unique relations predicted by the models for different datasets. As we can see, ART strongly outperforms its sampling baselines in predicting diverse relations.}
    \label{fig:unique_relations}
\end{figure*}

\begin{figure*}[ht]
    \centering
    \includegraphics[width=\linewidth]{figures/model_comparison_unseen.pdf}
    \caption{\textbf{Total number of unseen relation predictions of Vicuna and FlanT5 model variants.} The plots depict the total number of unique new relations predicted by the models, \ie the relations have been never seen during training using the VG train set. As we can see, ART significantly outperforms its baselines at predicting unseen relations. }
    \label{fig:unique_unseen_relations}
\end{figure*}
\end{comment}

\section{Computational cost and predictive perfomance}
In this section, we analyze both the computational characteristics and the predictive performance behavior of ART. We provide a breakdown of training and inference time, examine the trade-off between data usage and predictive performance, and explain the observed drop in R@k metrics due to biased relation distributions in evaluation datasets.

\subsection{Computational cost}
As depicted in \cref{tab:time_comp}, while ART incurs higher training costs due to adaptive sampling, it does not increase inference time, making it practical for real-world deployment. The added training complexity is offset by ART’s superior generalization, ensuring improved relation prediction without sacrificing efficiency during inference. This trade-off is crucial, as ART enhances mean Recall (mR) by prioritizing informative samples, ultimately leading to a more robust VRD model that generalizes well to unseen data.

\begin{table}[t]
\caption{\textbf{Comparison of training and inference time} on a single A100 GPU.}
\label{tab:time_comp}
\centering
\footnotesize  
\setlength{\tabcolsep}{3pt}  
\renewcommand{\arraystretch}{0.9}
\vspace{-0.5em}
\begin{tabularx}{\linewidth}{@{}X c c@{}}
\toprule
\textbf{Method} & \textbf{Train (hrs)} & \textbf{Inference (sec/Itr)} \\
\midrule
SGGs (Motifs, VTransE, VETO) & 18–22  & 0.07–0.075 \\
VLM (Random/Balanced)        & 32     & 0.45  \\
VLM (Adaptive)               & 96     & 0.45  \\
\bottomrule
\end{tabularx}
\vspace{-0.5em}
\end{table}

\subsection{Computational cost \vs performance trade-off}
\begin{figure}[h]
\centering
\includegraphics[width=0.95\linewidth]{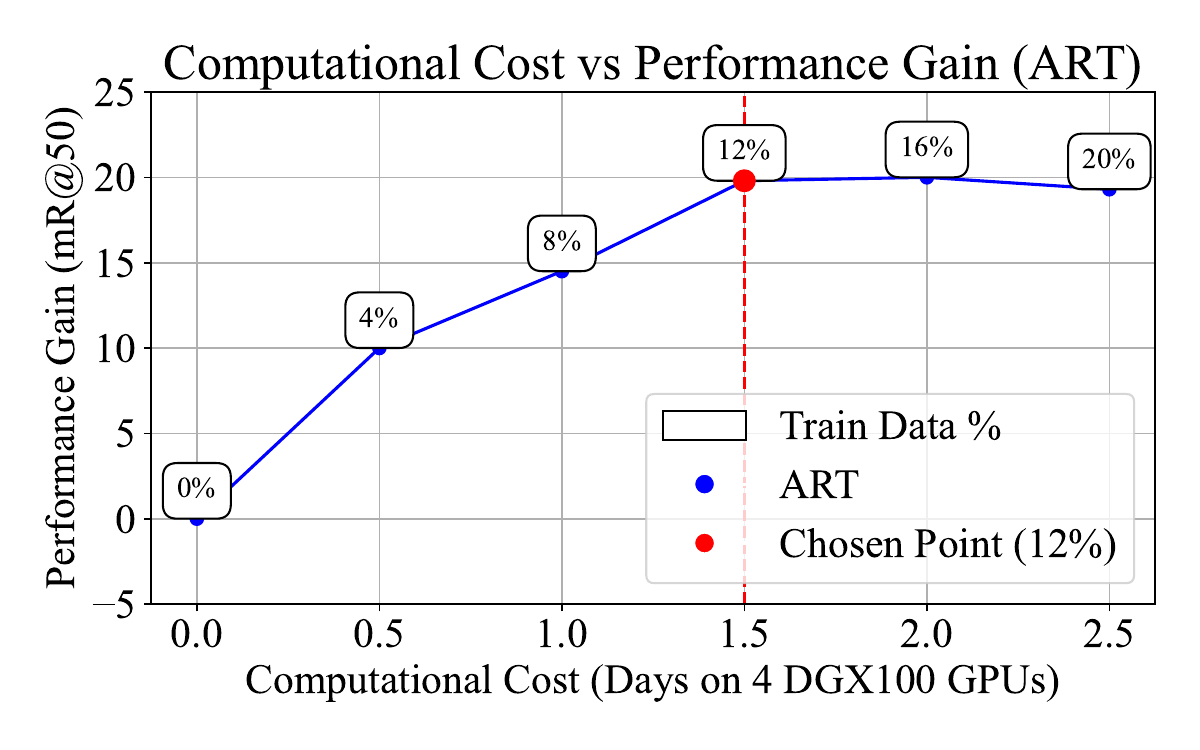}
\vspace{-1em}
\caption{\textbf{Trade-off between computational cost and predictive performance} as a function of training data usage.}
\label{fig:data_usage_2}
\vspace{-0.5em}
\end{figure}

As shown in \cref{fig:data_usage_2}, using just 12\% of the training data provides an excellent trade-off between computational cost and predictive performance. This setting achieves near-peak accuracy while requiring only 1.5 days of training on four DGX-A100 GPUs. Beyond this point, additional data yields diminishing returns.

Notably, the 0\% baseline incurs negligible computational cost but delivers limited predictive performance, whereas the 12\% configuration offers substantial gains at a reasonable expense. 

\subsection{On R@k and gR@k performance trade-offs}
While ART achieves strong generalization and diversity, it may show lower R@k and gR@k compared to random baselines in cases where ground-truth annotations are skewed toward frequent but semantically shallow relations. Random sampling tends to exploit this bias by favoring head-predicate predictions, leading to inflated recall scores without improving meaningful understanding. In contrast, ART explicitly counteracts this bias through balanced and adaptive sampling, resulting in more informative and diverse predictions. This is evidenced by the higher number of unique and unseen relations predicted by ART across datasets (\cref{fig:vicuna_predictions,fig:flan_predictions}) and illustrated qualitatively in the relation prediction examples (\cref{sec_qr}).

\section{Additional analysis: Adaptive \vs fixed thresholding} 
\label{app:thresholding}
As depicted in \cref{tab:thresholding}, we begin with fixed midpoints (2\textsuperscript{nd} row) for threshold values: 0.5 for entropy scores ($h_{\text{FN}}$, $t_{\text{FN}}$, $t_{\text{FP}}$) and 0.95 for similarity scores ($t_{\text{FP}}$). The higher similarity threshold accounts for the fact that similarity is computed on predicate phrases, not standalone predicates, resulting in generally higher values (\eg, \cref{fig:art}, instance \circled{B}{yellow}).  We then lower $t$ and increase $h$ from their midpoint values to explore fixed thresholding. However, all fixed-threshold variations yield lower mR than adaptive thresholding, highlighting the difficulty of selecting an optimal fixed threshold. In contrast, adaptive thresholding dynamically adjusts per predicate, ensuring optimal tuning of the VLM.
\begin{table}[t]
\centering
\caption{\textbf{Adaptive \vs fixed thresholding.}
$t_{\text{FP}}$: low-similarity FP threshold,
$t_{\text{FN}}$: low-entropy FN threshold,
$h_{\text{FN}}$: high-entropy FN threshold,
$h_{\text{TP}}$: high-entropy TP threshold. From the mid point threshold, we increase (higher-$h$) or decrease (lower-$t$) the respective thresholds to analyse the effect of fixed thresholds.}
\vspace{-0.5em}
\resizebox{\linewidth}{!}{%
\begin{tabular}{@{}lcccccc@{}}
\toprule
Fixed Thresholding & $t_{\operatorname{FP}}$ & $t_{\operatorname{FN}}$ & $h_{\operatorname{FN}}$ & $h_{\operatorname{TP}}$ & R@20/50 & mR@20/50 \\
\midrule
Lower-$t$ & 0.9 & 0.25 & 0.5 & 0.5 & \textbf{42.1/42.7} & 44.8/45.9 \\
Mid point & 0.95 & 0.5 & 0.5 & 0.5 & 37.2/37.4 & 43.2/44.3 \\
Higher-$h$ & 0.95 & 0.5 & 0.75 & 0.75 & 34.5/34.8 & 44.7/45.9 \\
Lower-$t$ Higher-$h$ & 0.9 & 0.25 & 0.75 & 0.75 & 34.5/34.9 & 44.7/45.9 \\
\midrule
Adaptive Thresholding & & & & & 41.1/41.4 & \textbf{46.4/47.7} \\
\bottomrule
\end{tabular}%
\vspace{-0.5em}
}

\label{tab:thresholding}
\end{table}

\section{Qualitative results}
\label{sec_qr}
Next, we present some qualitative results of ART on downstream segmentation, followed by a comparative analysis of relation predictions between ART and its baselines.

\subsection{ART-enhanced segmentation reasoning}
\begin{figure}[t]
    \centering
    \includegraphics[width=.95\linewidth]{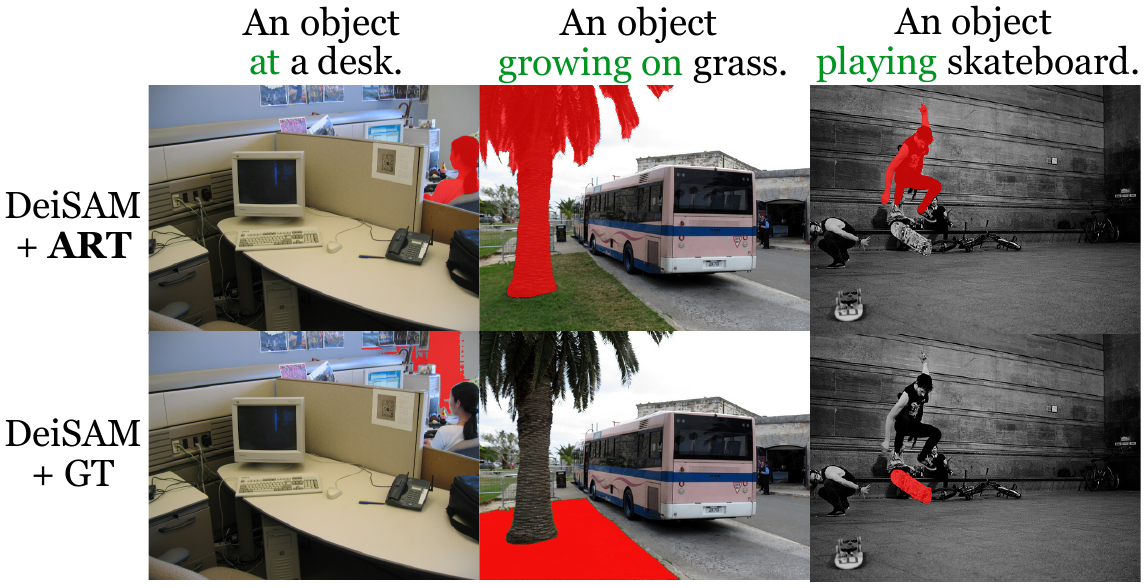}
    \vspace{-0.5em}
        \caption{\textbf{ART can be used to label missing annotations and predict new unseen predicates.} Segmentation results with textual prompts \emph{(top)} using DeiSAM~\cite{shindo2024deisam}, which segments objects via reasoning on scene graphs. ART successfully detects new relations and improves the segmentation quality, while ground-truth scene graphs fail to capture relations in the prompt.}
    \label{fig:deisam_result}
    \vspace{-0.5em}
\end{figure}

As shown in \cref{fig:deisam_result}, ART-enhanced scene graphs enable DeiSAM~\cite{shindo2024deisam} to produce higher-quality segmentations. While ground-truth scene graphs fail to capture the relations in the segmentation prompt, ART’s unseen relation prediction allows DeiSAM to accurately segment the referenced object in the deictic prompt.

\subsection{Comparative analysis of ART and its baselines}

\begin{figure*}[h]
    \centering
    \includegraphics[width=\linewidth]{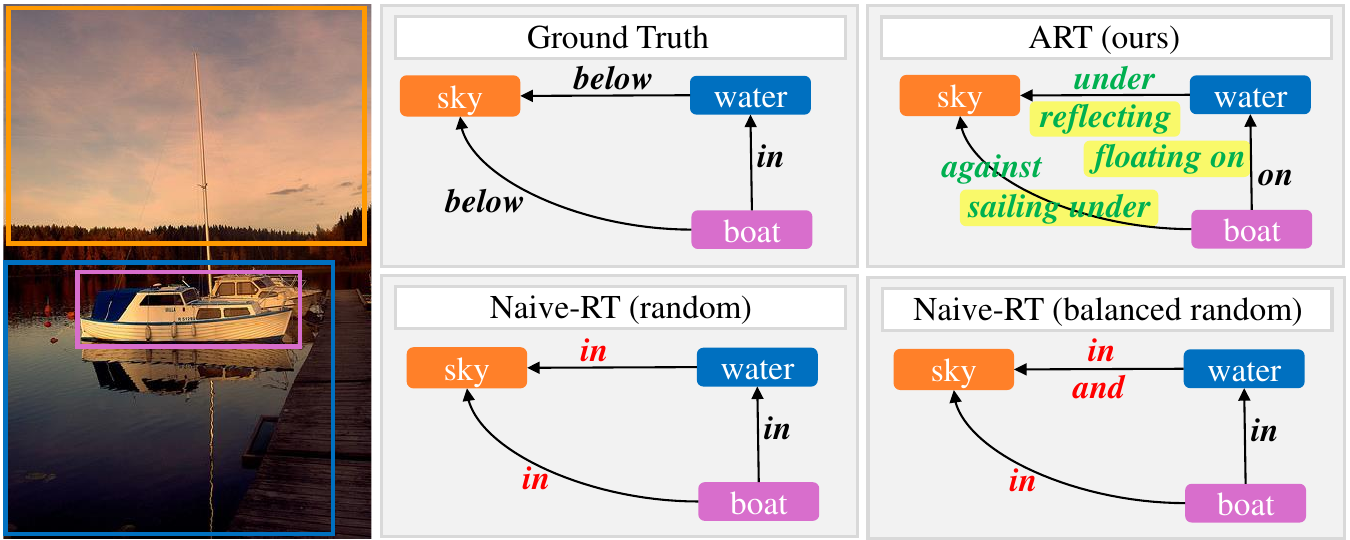}
    \caption{\textbf{Comparison of ART and its baselines on the GQA dataset.} ART predicts sensible spatial relations similar to the ground-truth annotation such as, \emph{water under sky}, while also identifying more informative relations than the ground truth, such as \emph{water reflecting sky}, \emph{boat floating on water}, and \emph{boat sailing under sky}. In contrast, both Naive-RT (random) and (balanced random) perform poorly. Informative relation predictions are highlighted in green, while those that are both informative and unseen are additionally highlighted in yellow. Incorrect predictions are marked in red.}
    \label{fig:reflecting_gqa}
    
\end{figure*}

\begin{figure*}[h]
    \centering
    \includegraphics[width=\linewidth]{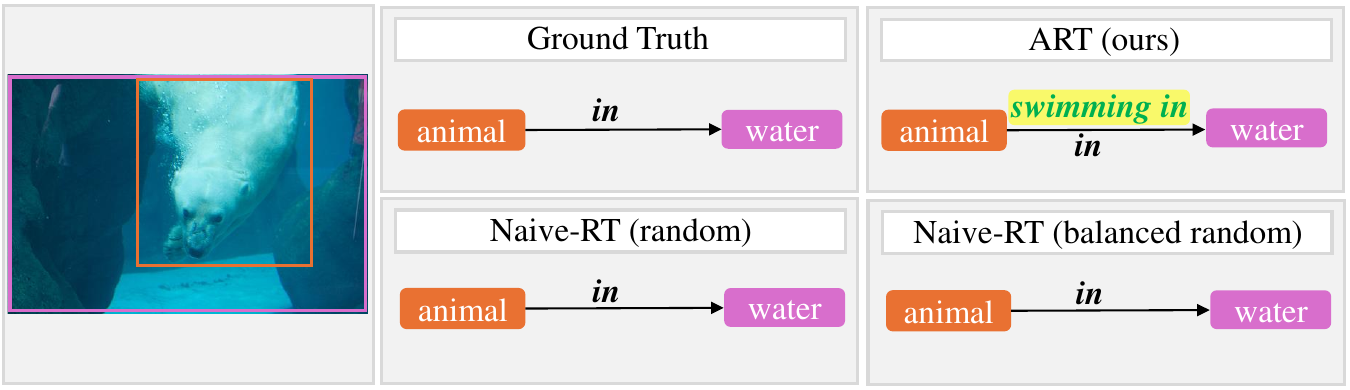}
    \caption{\textbf{Comparison of ART and its baselines on the GQA dataset.} The ground truth only provides a spatial relation \emph{in} between animal and water, while ART predicts the descriptive interaction \emph{swimming in}. Informative relation predictions are highlighted in green, while those that are both informative and unseen are additionally highlighted in yellow.}
    \label{fig:swimming_gqa}
    
\end{figure*}

\begin{figure*}[h]
    \centering
    \includegraphics[width=\linewidth]{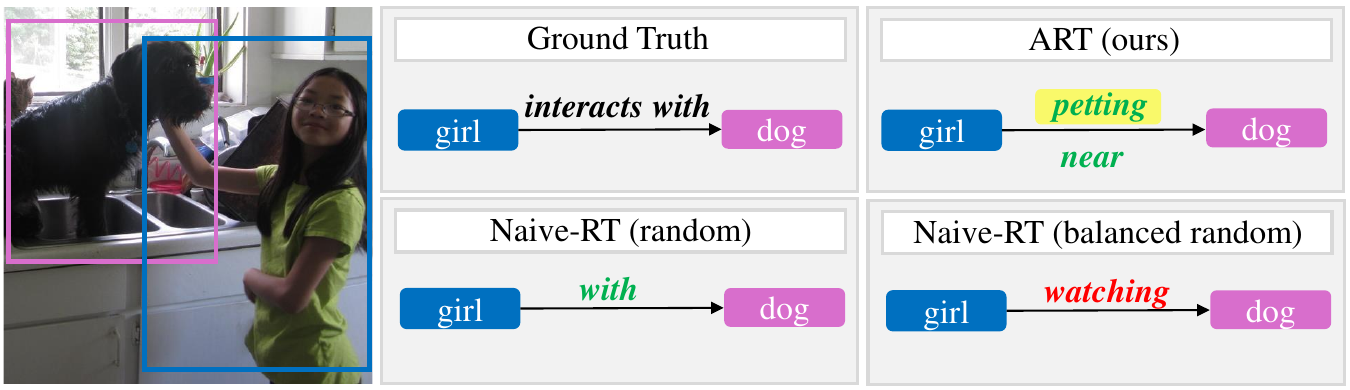}
    
    \caption{\textbf{Comparison of ART and its baselines on the OI-v4 dataset.} In contrast to the provided less informative ground-truth relation \emph{interacts with} in \emph{girl interacts with dog}, which raises the question ``What kind of interaction?", ART provides a much clearer interpretation that the interaction is \emph{petting}, \ie \emph{girl petting dog}, while also identifying the sensible spatial relation \emph{near}. In contrast, while Naive-RT (random) suggests the less meaningful relation \emph{with}, Naive-RT (balanced random) produces an entirely incorrect prediction. Informative relation predictions are highlighted in green, while those that are both informative and unseen are additionally highlighted in yellow. Incorrect predictions are marked in red.}
    \label{fig:petting_oiv4}
    
\end{figure*}

\begin{figure*}[h]
    \centering
    \includegraphics[width=\linewidth]{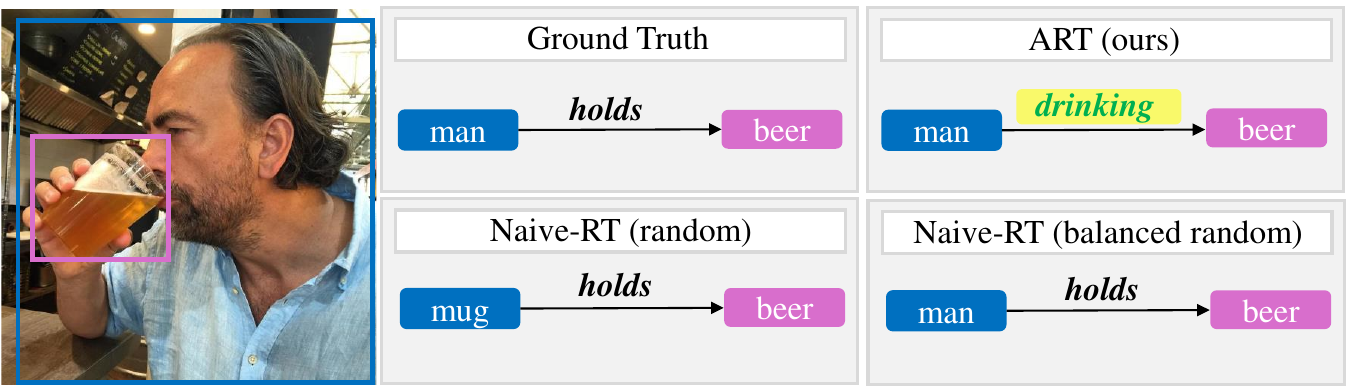}
    \caption{\textbf{Comparison of ART and its baselines on the OI-v4 dataset.} The ground truth relation \emph{holds} between \emph{man} and \emph{beer} leaves an open question ``What he intends to do with the beer?", while the prediction \emph{drinking} made by ART gives more context and the ongoing action. The Naive-RT baselines also predict the less descriptive relation \emph{holds}. Informative relation predictions are highlighted in green, while those that are both informative and unseen are additionally highlighted in yellow.}
    \label{fig:drinking}
    
\end{figure*}

\begin{figure*}[h]
    \centering
    \includegraphics[width=\linewidth]{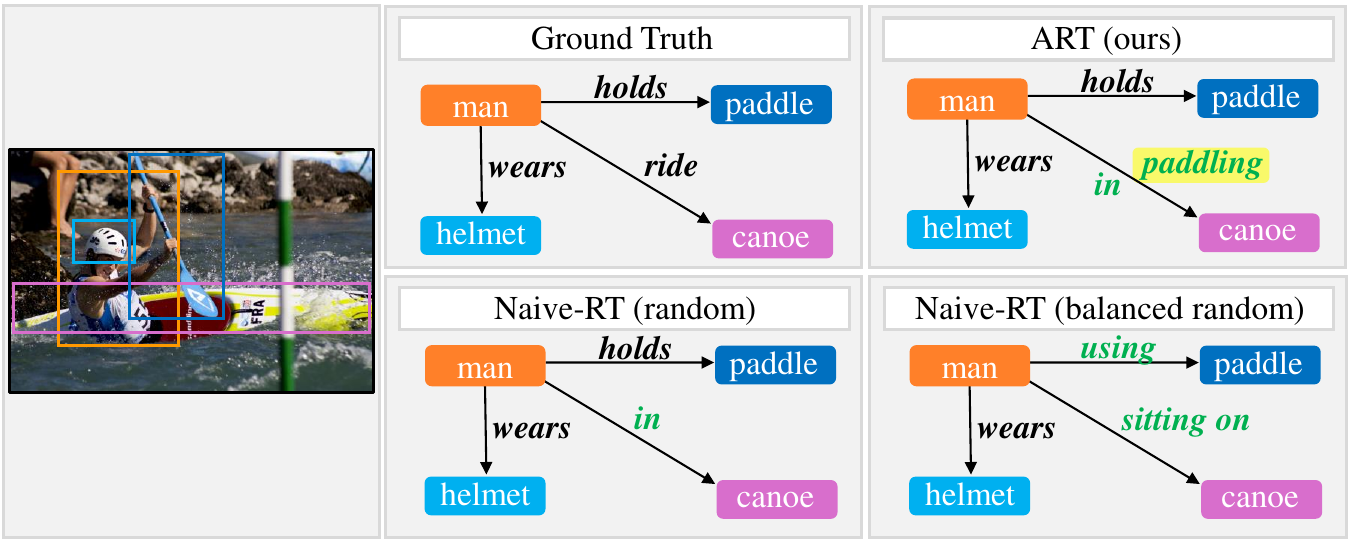}
    \caption{\textbf{Comparison of ART and its baselines on the OI-v6 dataset.} ART predicts the more informative relation \emph{paddling} between the man and canoe while also identifying sensible spatial relation \emph{in}. In contrast, although both Naive-RT (random) and (balanced random) make reasonable spatial predictions, they fail to clarify the action taking place between the man and the canoe. Informative relation predictions are highlighted in green, while those that are both informative and unseen are additionally highlighted in yellow.}
    \label{fig:paddling_oiv6}
    
\end{figure*}

\begin{figure*}[h]
    \centering
    \includegraphics[width=\linewidth]{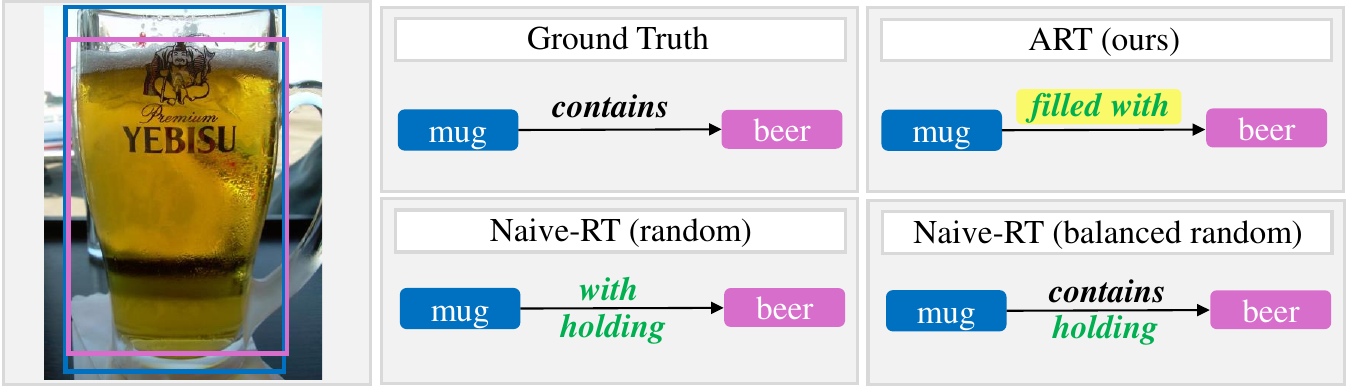}
    \caption{\textbf{Comparison of ART and its baselines on the OI-v6 dataset.} While the ground-truth relation \emph{contains} between the mug and beer merely indicates the presence of beer in the mug, the relation \emph{filled with}, predicted by ART, provides more detail by suggesting that the mug is full or nearly full of beer. The predictions \emph{with} and \emph{holding}, made by Naive-RT (random) and (balanced random) respectively, are reasonable but lack the level of descriptiveness conveyed by \emph{filled with}. Informative relation predictions are highlighted in green, while those that are both informative and unseen are additionally highlighted in yellow.}
    \label{fig:filled}
    
\end{figure*}

We compare the relationship predictions from the ART Vicuna model against its strongest baseline, Naive-RT (Naive Relation Tuning), which includes Naive-RT (balanced random) and Naive-RT (random), as well as the ground truth. Predictions are evaluated on the GQA, OI-v4, and OI-v6 test sets. 
Examples are shown in \cref{fig:reflecting_gqa,fig:swimming_gqa,fig:petting_oiv4,fig:drinking,fig:paddling_oiv6,fig:filled}.

Overall, we observe that ART not only identifies new relationships but also produces predictions that are more meaningful than existing ground-truth annotations. Predictions that are either similar to or more meaningful than the ground-truth annotation are highlighted in green,  while those that are both informative and unseen are additionally highlighted in yellow. Incorrect predictions are marked in red.

ART consistently outperforms its baselines across the GQA, OI-v4, and OI-v6 datasets by providing more meaningful and informative relationship predictions. For example, on the GQA dataset (see \cref{fig:reflecting_gqa}), ART predicts the sensible spatial relation \emph{under}  between water and sky and more detailed interactions such as \emph{water reflecting sky} and \emph{boat sailing under sky}, while Naive-RT (random) and (balanced random) perform poorly. \cref{fig:swimming_gqa} highlights that ART predicts the more descriptive interaction \emph{swimming in} between the animal and water, whereas the Naive-RT baselines, as well as the ground truth, only identify the spatial relation \emph{in}. On the OI-v4 dataset (see \cref{fig:petting_oiv4}), ART clarifies ambiguous ground-truth relations like \emph{interacts with}, which raises the question ``What kind of interaction?" by providing clarity that the interaction is \emph{petting} and also predicts the spatial relation \emph{near}, whereas Naive-RT baselines fail to provide clarity. Similarly, in \cref{fig:drinking}, the ground-truth relation \emph{holds} between man and beer raises the question, ``What does he intend to do with the beer?" This ambiguity is resolved by ART's prediction of the more specific relation \emph{drinking}. On the OI-v6 dataset (see \cref{fig:paddling_oiv6}), ART identifies the action \emph{paddling} between man and canoe, along with the spatial relation \emph{in}, outperforming the baselines, which lack specificity in describing the interaction. Another example from the OI-v6 dataset, shown in \cref{fig:filled}, once again shows that the ground-truth relation \emph{contains} between mug and beer is less detailed compared to ART's prediction of \emph{filled with}, which conveys that the mug is full or nearly full of beer.  The reasonable predictions \emph{with} and \emph{holding} made by Naive-RT (random) and (balanced random) are also less descriptive. 

Overall, the qualitative examples support the substantial quantitative gains reported in \cref{tab:results} of the main paper.

\end{document}